\documentclass[10pt,twocolumn,letterpaper]{article}

\usepackage{cvpr}
\usepackage{times}
\usepackage{epsfig}
\usepackage{graphicx}
\usepackage{amsmath}
\usepackage{amssymb}
\usepackage{gensymb}
\usepackage{multirow}
\usepackage{rotating}
\usepackage{fancyhdr}
\usepackage{latexsym}
\usepackage{url}
\usepackage{algorithm}
\usepackage{algorithmic}
\usepackage{bm}
\usepackage{stmaryrd}
\usepackage{dsfont}
\usepackage{epstopdf}
\usepackage{booktabs}
\usepackage{dashrule}

\AppendGraphicsExtensions{.tiff}

\usepackage{caption}
\usepackage{subcaption}

\usepackage[pagebackref=true,breaklinks=true,letterpaper=true,colorlinks,bookmarks=false]{hyperref}

\usepackage[sort, nocompress]{cite}

\cvprfinalcopy 

\def\cvprPaperID{7155} 

\ifcvprfinal\fi
\begin{document}
\addtolength{\baselineskip}{-0.25pt}
\title{Two-Stream FCNs to Balance Content and Style for Style Transfer}

\author{Duc Minh Vo\footnote{Corresponding author.}\\
SOKENDAI (Graduate University for Advanced Studies) \\
Tokyo, Japan \\
{\tt\small vmduc@nii.ac.jp}
\and
Akihiro Sugimoto\\
The National Institute of Informatics\\
Tokyo, Japan \\
{\tt\small sugimoto@nii.ac.jp}
}

\maketitle

\begin{abstract}

Style transfer is to render given image contents in given styles, and it has an important role in both computer vision fundamental research and industrial applications.
Following the success of deep learning based approaches, this problem has been re-launched recently, but still remains a difficult task because of trade-off between preserving contents and faithful rendering of styles.
Indeed, how well-balanced content and style are is crucial in evaluating the quality of stylized images.
In this paper, we propose an end-to-end two-stream Fully Convolutional Networks (FCNs) aiming at balancing the contributions of the content and the style in rendered images. Our proposed network consists of the encoder and decoder parts. 
The encoder part utilizes a FCN for content and a FCN for style where the two FCNs have feature injections and are independently trained to preserve the semantic content and to learn the faithful style representation in each. The semantic content feature and the style representation feature are then concatenated adaptively and fed into the decoder to generate style-transferred (stylized) images. In order to train our proposed network, we employ a loss network, the pre-trained VGG-16, to compute content loss and style loss, both of which are efficiently used for the feature injection as well as the feature concatenation. Our intensive experiments show that our proposed model generates more balanced stylized images in content and style than state-of-the-art methods. Moreover, our proposed network achieves efficiency in speed.


\end{abstract}

\section{Introduction}

\begin{figure*}[t!]
	\centering
	\includegraphics[width=1\linewidth]{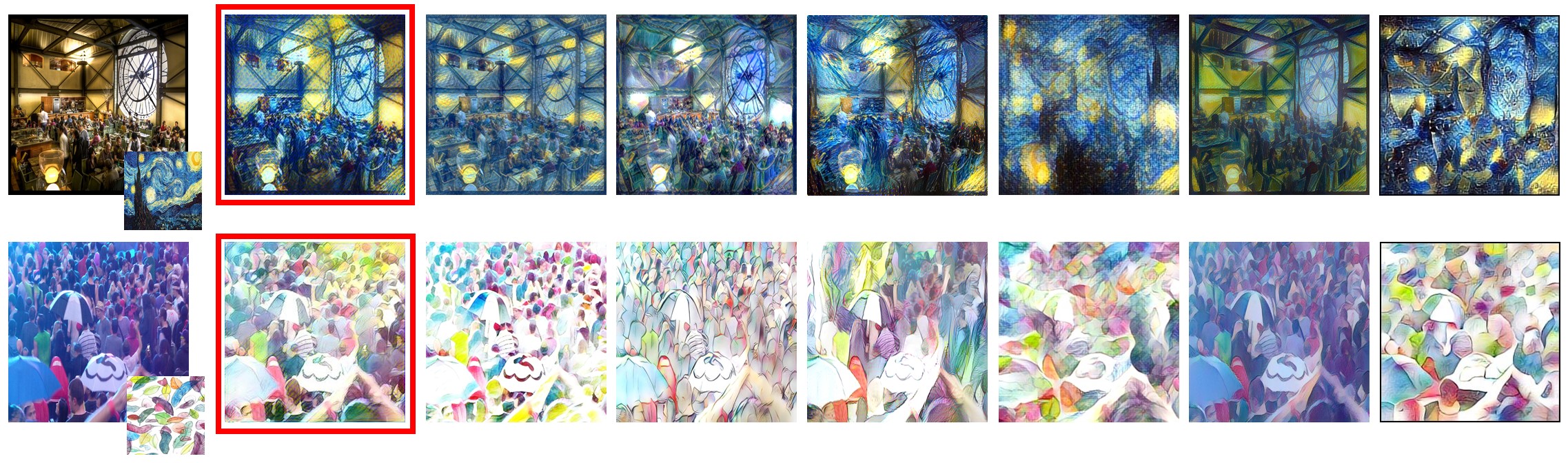}
	\caption{Example of stylized results. Left-most column: content image (large) and style image (small). From left to right: the stylized image by our method, Johnson+~\cite{Johnson2016Perceptual}, Huang+~\cite{huang2017adain}, and Gatys+~\cite{gatys2016image}, Sheng+~\cite{sheng2018avatar}, Chen+~\cite{Chen}, and Li+~\cite{Li2017Universal}. Our results surrounded with red rectangles are more balanced in content and style than the others. } 
	\label{fig:first_example}
\end{figure*}

\begin{figure}[tb]
	\centering
	\includegraphics[width=1\linewidth]{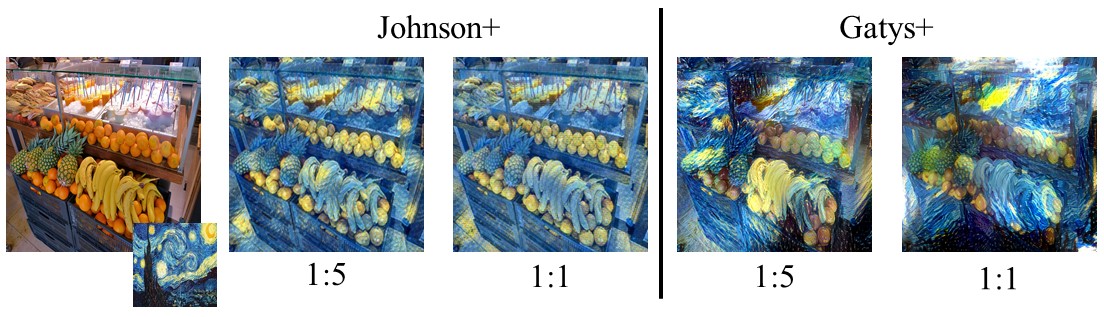}
    \vspace*{-1.0\baselineskip}
	\caption{Example of stylized results obtained by Johnson+~\cite{Johnson2016Perceptual} and Gatys+~\cite{gatys2016image} by changing the ratio of content and style from 1:5 to 1:1. Left-most column: content image (large) and style image (small). In each block, from left to right: the stylized image with various ratio of content and style.} 
	\label{fig:change_ratio}
\end{figure}

How New York looks like in ``The Starry Night'' by Vincent van Gogh is an interesting question and, at the same time, difficult to answer. 
In practice, re-painting a famous fine-art style takes much time and requires well-trained artists. 
Answering this question can be stated as the problem of migrating semantic content of one image to different styles, and it is called style transfer.

Style transfer is long-standing and has fallen into the image synthesis problem which is a fundamental research in computer vision.
Style transfer has its origin from non-photo-realistic rendering~\cite{Kyprianidis2013} and is closely related to texture synthesis and color transfer~\cite{Ashikhmin2001,Efros2001}. 
Along with the impressive progress of various tasks in computer vision using deep neural networks, this topic has recently been re-launched in both academy and industry. 
~\cite{gatys2016image} showed that the image representation derived from a Convolutional Neural Network (CNN) can be used to represent the semantic content of an image and the style, which opened up a new trend of CNN-based style transfer.

CNN-based approaches in style transfer fall into two categories~\cite{JingYFYS17}: Image-Optimisation-Based Online Neural Methods (IOB-NST) and Model-Optimisation-Based Offline Neural Methods (MOB-NST). 
The key idea of IOB-NST is to synthesis a stylized image by directly updating pixels in the image iteratively through the back-propagation. The IOB-NST such as~\cite{gatys2016image,luan2017deep,mechrez2017photo} starts with a noise image and iteratively updates the image by changing the distribution of noise along with the statistics of content and style until the defined loss function is minimized.
MOB-NST such as~\cite{azadi2018multi,Chen,huang2017adain,Johnson2016Perceptual,Dmytro2019Content,Li2017Universal,Park2019Arbitrary,sanakoyeu2018styleaware,sheng2018avatar,Wang2017CVPR}, on the other hand, first optimizes a generative model through iterations, and then renders the stylized image using a forward pass. In order to optimize the generative model, MOB-NST trains each feed-forward model for each specific style by using the gradient descent over a large dataset.
IOB-NST is known to produce better stylized results in quality than MOB-NST~\cite{JingYFYS17}, while MOB-NST has more efficiency in speed.

Although existing methods~\cite{azadi2018multi,Chen,gatys2016image,huang2017adain,Johnson2016Perceptual,Dmytro2019Content,Li2017Universal,luan2017deep,mechrez2017photo,Park2019Arbitrary,sanakoyeu2018styleaware,sheng2018avatar,Wang2017CVPR} show the capability of rendering image contents in different styles, generated stylized images are not always well balanced in content and style.
Such methods take care of either the content or the style, but not both, producing unbalanced stylized images.
IOB-NST is good at faithfully rendering the style while it tends to lose the content. 
MOB-NST, on the other hand, preserves more semantic content than the style. 
How to keep the balance between the content and the style in style transfer is a crucial issue to improve the quality of stylized images. 
This is because such balance is required in many applications; for instance, font transfer~\cite{zhang2018separating}, realistic photo transfer~\cite{li2018aclosed-form,mechrez2017photo}.
%
%
IOB-NST and MOB-NST have the capability of controlling the balance between the content and the style.  
Namely, they allow to manually change the ratio of content and style. 
However, changing the ratio do not guarantee that network parameters for stylized images changes as expected, 
meaning that the contributions of the content and the style in a stylized image are uncontrollable in reality. Fig.~\ref{fig:change_ratio} shows examples obtained by IOB-NST (Gatys+~\cite{gatys2016image}) and MOB-NST (Johnson+~\cite{Johnson2016Perceptual}) with various settings of contributions of the content and the style.
We can see although the ratio of content and style is significantly changed, the results do not change much.

Another important issue to address is the computational speed. 
Although MOB-NST such as~\cite{azadi2018multi,Chen,huang2017adain,Johnson2016Perceptual,Dmytro2019Content,Li2017Universal,Park2019Arbitrary,sanakoyeu2018styleaware,sheng2018avatar,Wang2017CVPR} are able to produce stylized images fast, they rely on a strong computational power. 
Therefore, either IOB-NST or MOB-NST is hard to apply to real-time applications.

We propose an end-to-end two-stream network for balancing the content and style in stylized images where contributions of the content and the style are adaptively taken into account. 
The encoder part of our network consists of the content stream and the style stream where the streams have different architectures. 
The two streams are connected by adaptive feature injection and independently trained to learn the semantic content or the style representation.
The content features and the style features are then combined in our proposed adaptive concatenation to ensure the balanced contribution of each stream. 
As the decoder part of our network, we use the feed-forward model to reduce the rendering time while we spend much time on learning like~\cite{Chen,huang2017adain,Johnson2016Perceptual,Li2017Universal,sheng2018avatar,Wang2017CVPR}. 
Unlike other methods that train a new model from the scratch for a yet unknown style, we fine-tune parameters from an existing model, allowing our network not only to accommodate fast training but also to easily adapt new styles.
Our experiments demonstrate that our method produces more balanced stylized images in both content and style than the state-of-the-art methods (Fig.~\ref{fig:first_example}).  They also show that our method runs about 22$\times$ faster than the state-of-the-art methods. We remark that our proposed model is trained for one style only, but it is easy to be fine-tuned to other styles incrementally with a low cost.

The rest of this paper is organized as follows. We briefly review and analyze related work in Section~\ref{related work}. 
Next, we analyze the semantic levels of image features for content and style in Section~\ref{semantic levels}.
Then, we present the detail of our proposed method in Section~\ref{proposed method}. Section~\ref{experimental setup} and Section~\ref{experimental results} discuss our experiments. Section~\ref{conclusion} draws the conclusion. We remark that this paper extends the work reported in~\cite{duc2018balancing}. Our main extensions in this paper are building a new network using both our proposed adaptive feature injection and concatenation, and adding more experiments.

\section{Related work} \label{related work}

\begin{figure*}[tb]
	\centering
	\includegraphics[width=0.8\linewidth]{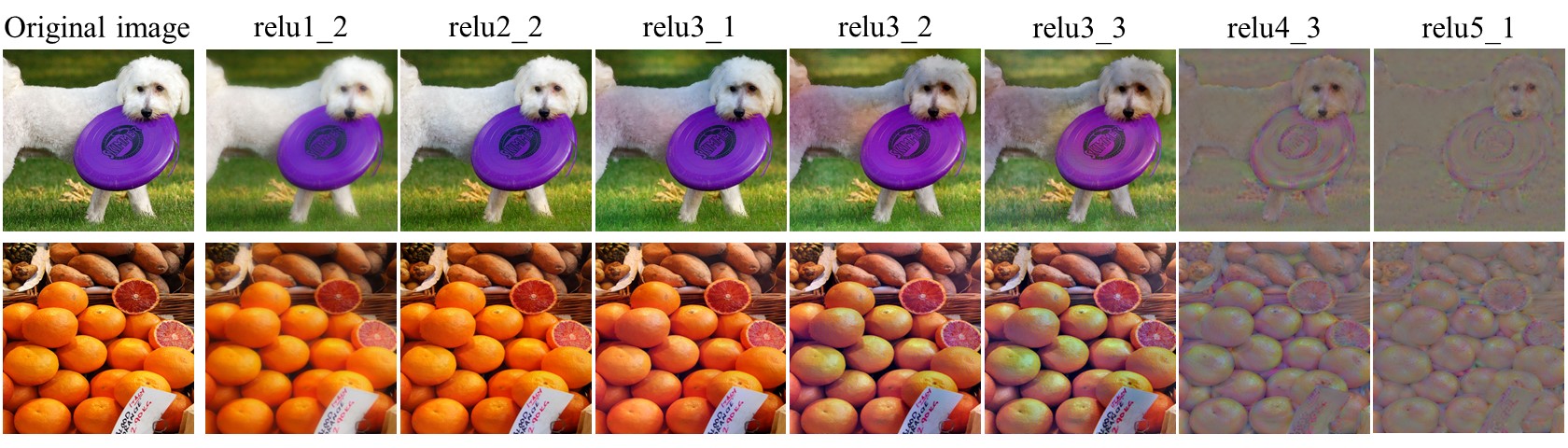}
	\caption{Examples of the feature reconstruction for several layers from the VGG-16 pre-trained network.} 
	\label{fig:content_reconstruction}
\end{figure*}

\begin{figure*}[tb]
	\centering
	\includegraphics[width=0.8\linewidth]{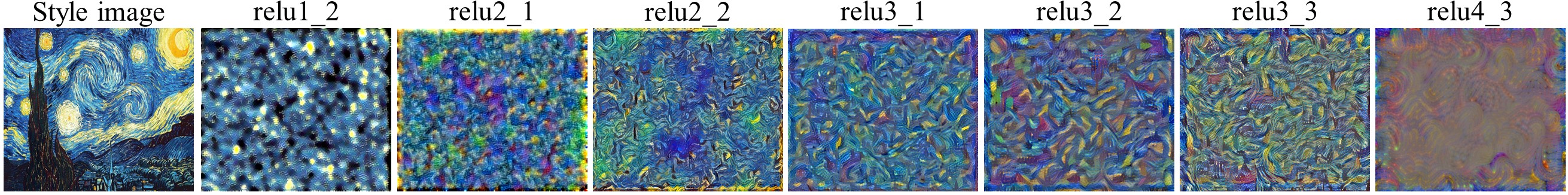}
	\caption{Examples of style image reconstruction for several layers from the VGG-16 pre-trained network.} 
	\label{fig:style_reconstruction}
\end{figure*}

\begin{figure*}[tb]
	\centering
	\includegraphics[width=0.7\linewidth]{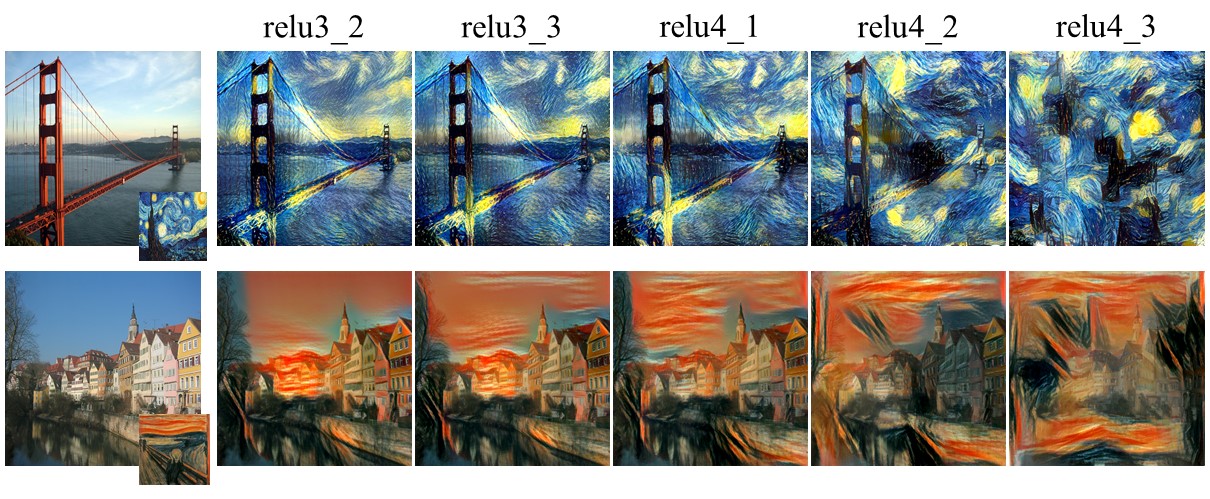}
	\caption{Examples of combination of content and style images from $relu3\_2$ to $relu4\_3$. Left-most column: content image (large) and style image (small), From left to right: the stylized images at different combination levels by Gatys+ \cite{gatys2016image} where the ratio of contributions of content and style is 1:1.} 
	\label{fig:matching_level}
\end{figure*}

Early work on style transfer was reported in the context of texture synthesis. Some methods there used histogram matching~\cite{Heeger} and/or non-parametric sampling~\cite{Ashikhmin2001,Efros2001}. These methods had limited results because they relied on hand-crafted low-level features and often failed in capturing features in semantic levels from the content and the style.

\cite{gatys2016image} for the first time proposed a method using CNNs and showed remarkable results. 
Their method trains CNNs to learn the semantic information from content images and matched it with the distribution of the style.
It starts from a randomly distributed noise image and iteratively updates the image to produce an image satisfying the semantic distribution of the content image and appearance statistics of the style.
During the iteration, the weighted sum of style loss and content loss is minimized.
As follow-up work of~\cite{gatys2016image}, \cite{luan2017deep} proposed a structure preservation method using Matting Laplacian for photo-realistic style transfer. 
\cite{mechrez2017photo} utilized the screened Poisson equation to make a stylized image more photo-realistic.
\cite{Li2017Laplacian} proposed a Laplacian loss that computes the Euclidean distance between the Laplacian filters responding to a content image and a stylized image in order to keep a fine structure of the content image. 
These approaches fall into the IOB-NST category, and all face with the computational speed problem.

\cite{Johnson2016Perceptual} and \cite{Ulyanov2016Texture}, on the other hand, took MOB-NST, proposing a feed-forward CNN and used the perceptual loss function for gradient-based optimization. 
The perceptual loss used there is similar to content and style loss in~\cite{gatys2016image}. 
Their models have only to pass the content image to a single forward network to produce a stylized image, which is fast.
Their two models are different only in the network architecture. 
\cite{Johnson2016Perceptual} follows the design of~\cite{Radford2015Unsupervised} with their modification of using residual blocks and fractionally strided convolutions while \cite{Ulyanov2016Texture} uses a multi-scale in their generator.
\cite{Wang2017CVPR} also utilized the feed-forward network, and they used multiple-generator to improve the quality of results. These methods are fast in generating stylized images, but they are capable of dealing with a single style only.

\cite{dumoulin2017learned} proposed a multi-style network that introduces shared-computation in many style images where they used instance normalization (IN)~\cite{UlyanovVL16} for balancing features from the content and from the style. They also proposed an improvement of IN to learn a different set of affine parameters for multi-styles in the batch way. 
However, their model can train a limited number of styles because the network capability is limited, meaning
that the number of styles to handle is limited. 
\cite{Chen} proposed a method that overcomes the limitation of the number of styles by using a patch-based method. 
Their method first extracts a set of patches from the content and style each, and then, 
for each content patch, the method finds its closest style patch and swaps their activation. 
In this way, their method transfers an unlimited number of styles; however, the cost for patch extraction and swapping increases the computational time significantly.
\cite{Li2017Universal} also proposed a method for multi-style transfer using feature transformations. They first employ pre-trained VGG-19 as their encoder to train an decoder for image reconstruction. Then, with fixing both encoder (VGG-19) and decoder, their model performs the style transfer through whitening and coloring transforms on a given content image and a style image. Though their method successfully solves the multi-style transfer, it still suffers from the computational cost and loses the content due to the feature transformations.

\cite{huang2017adain} and \cite{sheng2018avatar} proposed multi-style transfer models consisting of two CNN streams for content and style. 
\cite{huang2017adain} employed the pre-trained VGG-16 to extract content and style features and introduced Adaptive Instance Normalization (AIN) to make the mean and the variance of content features similar to those of style features. 
\cite{sheng2018avatar}, on the other hand, proposed AvatarNet which employed the pre-trained VGG-19 to extract the content and style features. 
These features are matched by using style-swap~\cite{Chen} or AIN~\cite{huang2017adain} before being fed into the decoder. 
Different from~\cite{huang2017adain}, their models have skip-connections from the style encoder to the decoder.
\cite{huang2017adain} and~\cite{sheng2018avatar}, however, used the same architecture for the content CNN and for the style CNN.
Having the same CNN architecture for the content and the style causes unavoidable unbalance between the content and the style because semantic levels extracted from the content and the style should not be the same in style transfer. 
Those models require expensive computational cost as well.
Furthermore, AIN~\cite{huang2017adain} assumes the standard distribution on pixel values of images, which is not always ensured in styles when normalizing data. 
In deed, AIN~\cite{huang2017adain} tends to produce a lot of artifacts; especially they are visible on flat surfaces~\cite{sanakoyeu2018styleaware}.
We remark that the skip-connection in AvatarNet~\cite{sheng2018avatar} weights the style contribution more, causing unbalance in stylized images.

Along with using Generative Adversarial Network (GAN)~\cite{ian2014generative} in image synthesis, several GAN-based models for style transfer are also proposed~\cite{azadi2018multi,Dmytro2019Content,Li2016Precomputed,sanakoyeu2018styleaware}. 
These models also optimize the network with a large number of content images during the training step, and thus fall in the MOB-NST category.
Though GAN-based models bring a promising approach to improve the quality of stylized images, their results, at this time, still are less impressive~\cite{JingYFYS17}. 
Furthermore, as in common with other GAN-based approaches, their training processes are also unstable.

Different from the methods above, we take into account the contributions of the content and the style through a two-stream feed-forward network to balance the content and the style in stylized images.
In particular, our proposed two-stream network is different from~\cite{huang2017adain,sheng2018avatar} in that our network has different depths in layer for the content and the style encoders to extract different semantic levels of the content and the style.
In addition, separating content and style enables our method easy to fine-tune to other styles with a cheaper computational cost (re-training time, required numbers of training images) than other models possessing only one encoder~\cite{Chen,Johnson2016Perceptual,Wang2017CVPR}. As a result, our method is able to easily deal with multi-styles.

\section{Semantic levels of image features for content and style} \label{semantic levels}

Along with the depth, CNN is known to extract different semantic levels of image features in layers.
As demonstrated in~\cite{gatys2016image,Johnson2016Perceptual}, features in early layers reflect colors, textures, and common patterns of images while those in latter layers preserve content and spatial structure of images. 
We, therefore, expect that the features in lower layers work as style features and those in higher layers do as content features. 
Using appropriate semantic levels of image features in style transfer is crucial.
We thus experimentally exploit the semantic levels of image features in VGG-16~\cite{SimonyanZ14a} to design suitable numbers of layers in designing our network to extract content and style features.
We remark that we refer \cite{gatys2016image,Johnson2016Perceptual} in which image reconstruction is learned using hidden features in CNN layers.

For the content image reconstruction, we randomly prepare 100 images. 
We then feed each of the 100 images into the VGG-16~\cite{SimonyanZ14a} pre-trained on object recognition using ImageNet dataset~\cite{RussakovskyDSKS15} without any fine-tuning and extract the features at each Rectified Linear Unit (ReLU)~\cite{Nair2010}. 
These features are employed to reconstruct original images using inverting technique~\cite{MahendranV15understanding}.
Hereafter, we use $reluX\_Y$ to mention a specific ReLU layer; see the definition of VGG-16~\cite{SimonyanZ14a} architecture for details.
Fig.~\ref{fig:content_reconstruction} shows some examples of image reconstruction at several layers.
We see that at low levels, i.e., from the $2$nd layer ($relu1\_2$) to the $5$th layer ($relu3\_1$), the reconstructed images are similar to the original image, meaning that these layers successfully keep colors, textures, and common patterns of images.
At higher levels, i.e., from the $6$th layer ($relu3\_2$) to the $10$th layer ($relu4\_3$), the reconstructed images preserve the content and spatial structure.
At even higher layers that start from the $11$th layer ($relu5\_1$), semantic features are gradually learned; the exact shape, on the other hand, is not preserved.

For the style image reconstruction, we use Adam optimization~\cite{Kingma2014} to find an image that minimizes the style reconstruction loss (proposed in~\cite{gatys2016image}). 
To obtain style reconstructed images, we start from a noise image and optimize the style loss as~\cite{gatys2016image} using the VGG-16 pre-trained on ImageNet. 
Fig.~\ref{fig:style_reconstruction} shows an example of the style image reconstruction.
We see that the style of image can be obtained until the $7$th layer ($relu3\_3$)

The above observation holds true for the images and the styles that we evaluated.
Combining the insight given by~\cite{gatys2016image,Johnson2016Perceptual}, we may thus conclude that the low-level layers reflect the style of the image while the high-level layers capture the content of the image. 
More precisely, from the $6$th layer ($relu3\_2$) to the $10$th layer ($relu4\_3$), the network is capable of appropriately capturing content information in the images. 
The style information, on the other hand, can be obtained from the $2$nd ($relu1\_2$) to the $7$th ($relu3\_3$) layers.

\cite{gatys2016image} pointed out that image content and style cannot be completely disentangled. 
This indicates that depending on the objective, we have to appropriately design the layer levels of content and style features for their combination. 
We thus further analyze effectiveness of the layers from the $6$th ($relu3\_2$) to the $10$th ($relu4\_3$) for content matching to determine the best one for combination.
We follow \cite{gatys2016image} to synthesize the stylized images where we set the contributions of content and style to be equal with each other.
To this end, we fix the style matching from the $2$nd ($relu1\_2$) to the $7$th ($relu3\_3$) layers, while performing the content matching at every single layer from the $6$th ($relu3\_2$) to the $10$th layers ($relu4\_3$).
Fig.~\ref{fig:matching_level} shows examples of stylized images having different layers in combination.
We see that the content matching at the $6$th and the $7$th layers ($relu3\_2$ and $relu3\_3$) 
is most reasonable to keep the balance of content and style in stylized images.

Using above observation, we design our network to fully exploit the characteristics of image features.
We choose the $6$th layer for content because it has a smaller number of parameters than the $7$th layer (it is faster to learn).
We choose the $4$th layer for style because it is neither too early in layer nor marginally different from the layer used for content.
In conclusion, we use the features at the $6$th layer ($relu3\_2$) for content and those at the $4$th layer ($relu2\_2$) for style.

\begin{figure*}[t!]
	\centering
	\includegraphics[width=1\linewidth]{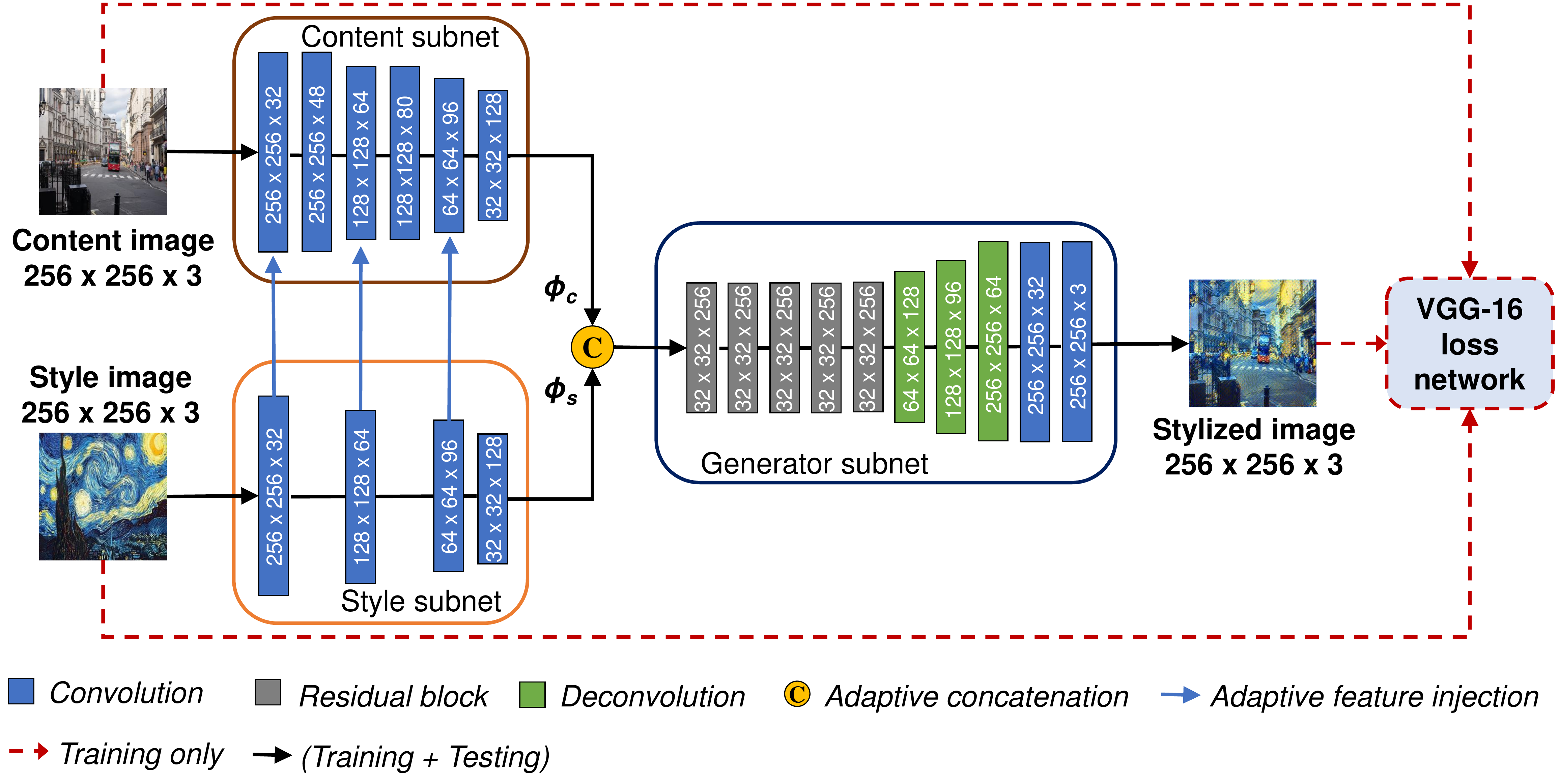}
    \vspace*{-0.5\baselineskip}
	\caption{Framework of our proposed method.  Our network consists of two encoders having different architectures and one decoder.  The loss network is used to train the encoders and the decoder.} 
	\label{fig:framework}
\end{figure*}

\section{Proposed method} \label{proposed method}
\subsection{Network design}

Our network follows end-to-end encoder-decoder architecture for rendering of the content in a given style~\cite{Chen,Johnson2016Perceptual,Wang2017CVPR}.
The network in~\cite{Chen,Johnson2016Perceptual,Wang2017CVPR} possesses only one encoder to extract the semantic content and style. 
This means that the extracted semantic level of the content and that of the style are the same.  When we stylize images, the role of the content should be different from that of the style because the content gives us what exist (object shapes and locations) in the rendered image and the style gives us the impression of the rendered image.
Accordingly, the semantic level used for the rending should be different depending on the content or the style.  Otherwise, unbalance between the content and style remains in stylized images.
We thus design a network having two encoders in which their architectures are different from each other to extract different semantic levels of the content and the style.
With the two encoders, our model treats the content and the style in different ways, allowing the network to be able to balance the roles of the content and the style better than the model having only one encoder.

Ideally, the network should be able to retain the semantics of the content as well as the statistics of the style as much as possible. 
The semantic content and style of an image are captured at different layers in the network (see ~\cite{gatys2016image,Johnson2016Perceptual} and Section~\ref{semantic levels}): the network obtains the style at low-level layers in depth while high-level layers become more sensitive to the actual content of the image.
We thus design the encoders with different depths to retain useful information from both the content and the style. 
Namely, we design a deep encoder for the content and a shallow encoder for the style.
Moreover, in order to reflect features extracted from the style at low-level to those from the content, we employ the feature injection via the skip-connection technique from the shallow encoder to the deep one.
Because the content feature and the style feature are extracted at different levels in the network, they have different characteristics. We thus introduce an effective concatenation to enhance the contribution of these features for good performances instead of implementing their simple ones.

\subsection{Network architecture} \label{network architecture}

Our proposed network consists of three Fully Convolutional Network (FCNs): two encoders and one decoder (Fig.~\ref{fig:framework}). 
The two encoders are a deep network, the content subnet, to extract content feature $\phi_{\rm c}$ from a content image, and a shallow network, the style subnet, to extract style feature $\phi_{\rm s}$ from a style image.
The feature injection is employed between the content subnet and the style subnet using the balance weight (cf. Section~\ref{adaptive layer}).
This balance weight is also used to adaptively concatenate the features $\phi_{\rm c}$ and $\phi_{\rm s}$ at the top of content and style subnet before being fed into a deep network, the generator subnet, to produce a stylized image. 
We employ the VGG-16 model~\cite{SimonyanZ14a} as the loss network in the training phase.

Our network receives the content and style images where each image is with the size of $n \times n \times 3$ ($n$ is the size of image, 3 are for RGB channels), and synthesizes an stylized image of $n \times n \times 3$. In the training phase, we use the images of $256 \times 256 \times 3$ ($n$ = 256). Although we train the network on images with the size of $256 \times 256 \times 3$, the network can accept any size of images in testing ($n$ can be 64, 128, 256, or 512). We remark that the size of the content image and that of the style image have to be the same to ensure the consistency of the feature size when injecting and concatenating the content and the style features.

\subsubsection{Content subnet}
The content subnet is a stack of six convolution layers with the filter size of $3 \times 3$, and the padding size of $1 \times 1$. We use the stride of $2 \times 2$ at the third, the fifth, and the sixth layers to reduce the size of feature maps and the stride of $1 \times 1$ at the other layers. The numbers of the output channels are 32, 48, 64, 80, 96, and 128, respectively. Each convolution layer is followed by a spatial instance normalization (IN) layer~\cite{UlyanovVL16} 
and a Rectified Linear Unit (ReLU) layer~\cite{Nair2010}. In order to avoid the border artifacts caused by convolution, the reflection-padding is used instead of the zero-padding similarly to~\cite{dumoulin2017learned}.

\subsubsection{Style subnet}
The style subnet, which has four convolution layers, is shallow network (more precisely, shallower than the content subnet). All convolution layers have the filter size of $3 \times 3$, the reflection-padding of $1 \times 1$, and the stride of $2 \times 2$, except for the first layer that employs the stride of $1 \times 1$. The numbers of the output channels are 32, 64, 96, and 128, respectively. Similarly to the content subnet, each convolution layer is also followed by an IN layer~\cite{UlyanovVL16} and a ReLU layer~\cite{Nair2010}.

We employ feature injection from the feature $\phi_{\rm s}^q$ at the $q$-th layer in the style subnet to those $\phi_{\rm c}^p$ at the $p$-th layer in the content subnet, the size of whose feature map is the same (Table~\ref{tab:encoder architecture}). To take into account the contributions of $\phi_{\rm s}^{q}$ and $\phi_{\rm c}^{p}$, we introduce the adaptive feature injection with the balance weight (cf. Section~\ref{adaptive layer}).

\begin{table}[tb]
\centering
\caption{Architecture of our encoders. The arrow ($\leftarrow$) indicates the adaptive feature injection.} 
\label{tab:encoder architecture}
\vspace*{-0.5\baselineskip}
\resizebox{1\linewidth}{!}{

\begin{tabular}{@{\hspace*{0.35em}}c@{\hspace*{0.35em}}c@{\hspace*{0.35em}}c@{\hspace*{0.35em}}c@{\hspace*{0.35em}}c@{\hspace*{0.35em}}c@{\hspace*{0.35em}}c@{\hspace*{0.35em}}c@{\hspace*{0.35em}}c}
\toprule[1pt]\midrule[0.3pt]
\multicolumn{3}{c}{\textbf{Content subnet}} & & & & \multicolumn{3}{c}{\textbf{Style subnet}}\\
\cmidrule{1-3} \cmidrule{7-9}
{\textbf{No}} & {\textbf{Layer}} & {\textbf{Output channel}} & & & & {\textbf{No}} & {\textbf{Layer}} & {\textbf{Output channel}} \\ 
\midrule
0 & Content image & 3 & & & & 0 & Style image & 3 \\
1 & Convolution & 32 & & & & 1 & Convolution & 32\\
2 & Instance normalization & 32 & & & & 2 & Instance normalization & 32 \\
3 & ReLU & 32 & & $\leftarrow$ & & 3 & ReLU & 32\\
4 & Convolution & 48 \\
5 & Instance normalization & 48 \\
6 & ReLU & 48 \\
7 & Convolution & 64 & & & & 4 & Convolution & 64\\
8 & Instance normalization & 64 & & & & 5 & Instance normalization & 64\\
9 & ReLU & 64 & & $\leftarrow$ & & 6 & ReLU & 64\\
10 & Convolution & 80 \\
11 & Instance normalization & 80 \\
12 & ReLU & 80 \\
13 & Convolution & 96 & & & & 7 & Convolution & 96\\
14 & Instance normalization & 96 & & & & 8 & Instance normalization & 96 \\
15 & ReLU & 96 & & $\leftarrow$ & & 9 & ReLU & 96\\
16 & Convolution & 128 & & & & 10 & Convolution & 128\\
17 & Instance normalization & 128 & & & & 11 & Instance normalization & 128 \\
18 & ReLU & 128 & & & & 12 & ReLU & 128\\
\midrule[0.3pt]\bottomrule[1pt]
\end{tabular}
}
\end{table}

\subsubsection{Generator subnet}
The generator subnet consists of five residual blocks, three deconvolution layers, and two convolution layers in this order. 

\cite{Johnson2016Perceptual} argues that the residual block can enrich the information involved in the input feature. We, therefore, use residual blocks to increase the impact of the balance weight in the concatenated feature. Similarly to~\cite{Johnson2016Perceptual}, we use five residual blocks outputting 256 channels, where each of them has two convolution layers with the filter size of $3 \times 3$, the reflection-padding of $1 \times 1$, the stride of $1 \times 1$, and a summation layer as in~\cite{He2015}. All convolution layers are followed by an IN layer~\cite{UlyanovVL16} (we use it to replace the batch normalization~\cite{Ioffe2015} in the original architecture~\cite{He2015}) and a ReLU layer~\cite{Nair2010}. 

To upscale the feature map, we employ three deconvolution layers with the same filter size of $3 \times 3$, the reflection-padding of $1 \times 1$, and the stride of $2 \times 2$, outputting 128, 96, and 64 channels, respectively. 

In order to eliminate the affect of the convolution stride, we use two convolution layers which have the filter size of $1 \times 1$, the padding of $0 \times 0$, and the stride of $1 \times 1$, outputting 32 and 3 channels. All deconvolution layers and convolution layers are followed by an IN layer~\cite{UlyanovVL16} and a ReLU layer~\cite{Nair2010}, except for the last convolution layer that uses the $\tanh$ activation to guarantee that the range of the output can be normalized to be $[0, 255]$.

\subsection{Loss function} \label{loss function}
We employ two loss functions for content loss and style loss, which are computed from layers of the loss network. The content loss $\mathcal{L}_{\rm c}$ computes the similarity of high-level features between the content image and the stylized image. The style loss $\mathcal{L}_{\rm s}$, on the other hand, computes the similarity of low-level features between the style image and the stylized image. 

The overall loss is a weighted sum of the content loss and the style loss: 
\begin{equation}
\label{eq:total-loss}
\mathcal{L}(\hat{y},y_{\rm c},y_{\rm s})=\alpha\mathcal{L}_{\rm c}(\hat{y},y_{\rm c})+(1-\alpha)\mathcal{L}_{\rm s}(\hat{y},y_{\rm s}),
\end{equation}
where $y_{\rm c}, y_{\rm s}$, and $\hat{y}$ denote the content image, the style, and the stylized image, respectively. 
$\alpha$ is the combination weight (we set $\alpha=0.5$ in our experiments to equally weight these two loss functions).

We obtain the content loss at $M$ layers as follows: 
\begin{equation}\label{eq:content_loss2}
       \mathcal{L}_{\rm c}(\hat{y}, y_{\rm c})=\frac{1}{M}\sum_{k\in{M}}\frac{1}{C_k \times H_k \times W_k}\lVert{\Phi_k(\hat{y})-\Phi_k(y_{\rm c})}\lVert_2,
\end{equation}
where $\Phi_k(\cdot)$ denotes the normalized feature map at the $k$-th layer, which has $C_k \times H_k \times W_k$ elements. 
The range of $\mathcal{L}_{\rm c}$ is $[0, 1]$.

The style loss is computed at $N$ layers as follows: 
\begin{equation}\label{eq:style_loss2}
       \mathcal{L}_{\rm s}(\hat{y}, y_{\rm s})=\frac{1}{N}\sum_{k\in{N}}\lVert{G(\Phi_k(\hat{y}))-G(\Phi_k(y_{\rm s}))}\lVert_F,
\end{equation}
where $\lVert{\cdot}\lVert_F$ denotes the Frobenius norm~\cite{Horn2012}. $G(\Phi_k(\cdot))$ is the Gram matrix~\cite{Horn2012} of the normalized feature map at the $k$-th layer. The Gram matrix $G_{C_k \times C_k}$ has elements $G_{ij} = \langle{\upsilon_{i}, \upsilon_{j}}\rangle$ where $\upsilon_{i}, \upsilon_{j}$ are features at the $i$-th and the $j$-th channels respectively of the feature map $\Phi_k(\cdot)$.
The range of $\mathcal{L}_{\rm s}$ is $[0, 1]$.

\subsection{Adaptive feature injection and concatenation} \label{adaptive layer}

In our network, we employ the feature injection between the content features and the style features.
We also concatenate them to feed into the generator subnet. 
To weight the contributions of the content features and the style features, we introduce the balance weight $\gamma$.
This balance weight is adaptively updated during the training so that it retains the balance between the content and the style in stylized images.

At the $t$-th iteration in training phase, $\gamma_{\rm t}$ is computed as follows:
\begin{equation}\label{eq:balance_weight}
    \gamma_{\rm t} = \frac{\mathcal{L}_{\rm s}(t)}{\mathcal{L}_{\rm s}(t) + \mathcal{L}_{\rm c}(t)},  
\end{equation}
where $\mathcal{L}_{\rm s}(t)$ and $\mathcal{L}_{\rm c}(t)$ are the style loss and the content loss at the $t$-th iteration in the training phase.
To restrict the fluctuation of the balance weight, we compute $\gamma$ at every non-overlapping $T$ iterations and use it for the next $T$ iterations:
\begin{equation}\label{eq:practical_balance_weight}
    \gamma = \frac{1}{T}\sum_{t=1}^{T}{\gamma_t}.
\end{equation}

Using $\gamma$, we sum up the content feature at the $p$-th layer $\phi_{\rm c}^{p}$ and the style feature at the $q$-th layer $\phi_{\rm s}^{q}$ for the feature in adaptive feature injection as follows:
\begin{equation}\label{eq:skip-connection}
    \phi^{pq} = (\gamma \times \phi_{\rm c}^{p}) + ((1-\gamma)\times\phi_{\rm s}^{q}).
\end{equation}

Similarly, we concatenate the content feature $\phi_{\rm c}$ and the style feature $\phi_{\rm s}$ in the adaptive concatenation as follows:
\begin{equation}\label{eq:concatenation}
    \phi = (\gamma \times \phi_{\rm c}) \oplus ((1-\gamma)\times\phi_{\rm s}).
\end{equation}

The learned balance weight $\gamma$ ensures the balance of the contributions of the content feature and the style feature in both feature injection and concatenation layers. For example, when $\mathcal{L}_{\rm s}$ is smaller than $\mathcal{L}_{\rm c}$ (meaning $\gamma\leq 0.5$ in Eq.~(\ref{eq:practical_balance_weight})), the contribution of style feature is increased in the next iterations, and vice verse. Moreover, the learned balance weight $\gamma$ is more advantageous than the fixed balance weight that does not concern the balance of losses.


In order to explicitly control the contribution ratio of the content and the style, we manually set the expected contribution ratio in the loss function, and then introduce the learnable weight that allows us to change stylized images as we expect.
The combination weight $\alpha$ takes the former role while the learnable balance weight $\gamma$ does the latter role.
In other words, in our method, $\alpha$ sets an expected contribution ratio of content and style in stylized images through the loss function while $\gamma$ controls the learning direction of the network during the training to achieve the contribution ratio specified by $\alpha$.
In our experiments where we set $\alpha=0.5$, we see that $\gamma$ works for the equal contribution ratio of the content and the style as expected (see Sections \ref{qualitative evaluation} and \ref{quantitative evaluation} for details).
We remark that $\alpha$ and $\gamma$ together play the role of the indicator for how much the content and the style are emphasized in obtained stylized images.


\section{Experimental setup} \label{experimental setup}

\subsection{Dataset and compared methods}
\subsubsection{Dataset}
\begin{figure*}[tb]
	\centering
	\includegraphics[width=0.7\linewidth]{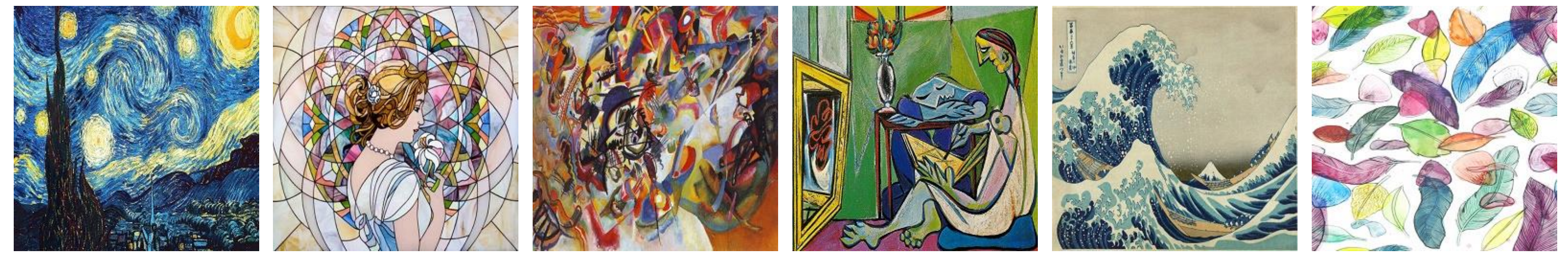}
	\caption{Styles used in experiments. From left to right: Starry Night, Mosaic, Composition VII, La Muse, The Wave, and Feathers.} 
	\label{fig:styles_in_experiment}
\end{figure*}

We used in our experiments, images in the MS-COCO 2014 dataset~\cite{Lin2014} as our content images, and six famous paintings widely used in style transfer~\cite{gatys2016image,huang2017adain,Johnson2016Perceptual}, as our style images (cf. Fig.~\ref{fig:styles_in_experiment}).

We used the MS-COCO 2014 training set for our training, and we randomly selected 20 images from the MS-COCO 2014 validation set for our validation. 
In the testing phase, on the other hand, we randomly selected 50 images from MS-COCO 2014 validation (different ones from the 20 images used in our validation). 
 
\subsubsection{Compared methods}
We compared our method with SOTA methods: Gatys+~\cite{gatys2016image}, Johnson+~\cite{Johnson2016Perceptual}, Huang+~\cite{huang2017adain}, Sheng+~\cite{sheng2018avatar}, Chen+~\cite{Chen}, and Li+~\cite{Li2017Universal}. We note that Gatys+ is based on IOB-NST and the others are on MOB-NST. For Gatys+, we used the re-implementation version 
by J. Johnson\footnote{https://github.com/jcjohnson/neural-style}. 
For the others, we used publicly available source codes with parameters recommended by the authors (Johnson+\footnote{https://github.com/jcjohnson/fast-neural-style}, Huang+\footnote{https://github.com/xunhuang1995/AdaIN-style}, Sheng+\footnote{https://github.com/LucasSheng/avatar-net}, Chen+\footnote{https://github.com/rtqichen/style-swap}, Li+\footnote{https://github.com/Yijunmaverick/UniversalStyleTransfer}). We remark that we set 1000 iterations for Gatys+.

\subsection{Implementation details}
\subsubsection{Implementation setup}
We implemented our method in PyTorch\footnote{https://pytorch.org/}. We used the instance incremental learning strategy for dealing with multiple styles. We conducted all experiments using a PC with CPU core i7 3.7 GHz, 12 GB of RAM, and GTX 770 GPU (4 GB of VRAM).

We performed the adaptive feature injection from layers $q=3,6,9$ in the style subnet to layers $p=3,9,15$ in the content subnet, respectively (Table~\ref{tab:encoder architecture}).
We adopted the VGG-16 model~\cite{SimonyanZ14a} pre-trained on the ImageNet~\cite{RussakovskyDSKS15} as the loss network without any fine-tuning. All layers after $relu4\_3$ layer were dropped. We obtained the content loss at $M = 1$ layer, e.g., $relu4\_3$, and the style loss at $N = 3$ layers, e.g., $relu1\_2$, $relu2\_2$, and $relu3\_3$ ($M$ and $N$ are defined in Section~\ref{loss function}).

\subsubsection{Training the model} \label{training model}

Our method addresses a one-style model to reduce computational time. 
For training a new yet unknown style, we fine-tune parameters from an existing model.
With this learning strategy, our method can easily adapt a new style with a lower cost than existing work~\cite{gatys2016image,huang2017adain,Johnson2016Perceptual,Wang2017CVPR}. 
Moreover, the fine-tuning learning enables our method to deal with an unlimited number of styles fast 
unlike existing methods such as~\cite{Chen,dumoulin2017learned}.

We first trained an initial model on the Starry Night style and then incrementally fine-tuned on the other styles one by one. We trained the network on the Starry Night style with a batch size of 2 for 80k iterations corresponding to 2 epochs. The balance weight $\gamma$ in Eq.~(\ref{eq:practical_balance_weight}) is re-computed at every $T=500$ iterations. All the training and validation images are resized to $256 \times 256$. To train the model, we used the Adam optimizer~\cite{Kingma2014} with the learning rate of $10^{-3}$, the moments $\beta_1=0.9$ and $\beta_2=0.999$, and the division from zero parameter $\epsilon=10^{-8}$. We did not use the learning rate decay and the weight decay.

For the initial model, we trained all subnets simultaneously with independently updating the weight of each subnet. Validation was performed at every 100 iterations during the training process. When observing the content loss and the style loss on the validation set, if any loss function raises the overfitting problem, we stopped updating the weight of the corresponding subnet.

We incrementally fine-tuned the initial model to the other styles one by one. 2000 images in the MS-COCO 2014 training set~\cite{Lin2014} were randomly selected as content images for training.
The network was trained for 1000 iterations with the batch size of 2. The Adam optimizer~\cite{Kingma2014} was also used with the same parameters as the training of the initial model. The balance weight $\gamma$ in Eq.~(\ref{eq:practical_balance_weight}) was re-computed at every $T = 50$ iterations. The loss-based training technique was also applied to avoid overfitting, where the validation was performed at every 50 iterations.

\subsection{Evaluation metric} 

In order to evaluate the quality of synthesized images, most previous work employed user studies although they are subjective and have ambiguity in evaluation.
We, on the other hand, evaluate stylized images by quantifying the content and style losses.
Intuitively, when the total loss is sufficiently small, we may say that the overall quality of stylized images is good.
Furthermore, the quality of stylized images also depends on how the content and the style are reflected in them.
We have to consider these two factors in evaluating the quality of stylized images.
Since the contributions of the content and the style are controlled by $\alpha$ (set in advance) in our method, we may see if a synthesized image is good in quality by evaluating (i) whether its total loss is sufficiently small, and (ii) whether the ratio between its content and style losses consistently agrees with the pre-set contribution ratio (i.e., combination weight $\alpha$) between the content and the style.
We thus introduce a metric to evaluate the quality of synthesized images using these two criteria. 
We remind that we set $\alpha=0.5$ (for simplicity) in our experiments to see the content and style losses converge to almost the same values.

For each pair of content image $c$ and style image $s$, we compute content loss ${\mathcal L}_{\rm c}$ and style loss ${\mathcal L}_{\rm s}$.  
In the 2D plane whose coordinate system is defined by content loss and style loss, the criterion (i) can be measured using
the distance between the origin and $({\mathcal L}_{\rm c},{\mathcal L}_{\rm s})$.
The criterion (ii), on the other hand, can be measured by evaluating how close $({\mathcal L}_{\rm c},{\mathcal L}_{\rm s})$ is to the line of ``content loss''$=$``style loss'' (called the balanced axis hereafter).

We assume that we have $K$ stylized images.  We normalize content loss and style loss for each stylized image over $K$ images:
\begin{equation}\label{eq:norm_content_loss}
    \widetilde{\mathcal L_{\rm c}} = \frac{1}{1 + \exp{(\frac{
    {\mathcal L}_{\rm c}-\overline{{\mathcal L}_{\rm c} }}
    {\sigma _{\rm c} } ) } }, \quad
%
    \widetilde{\mathcal L_{\rm s}} = \frac{1}{1 + \exp{(\frac{
    {\mathcal L}_{\rm s}-\overline{{\mathcal L}_{\rm s}}}
    {\sigma _{\rm s} } ) } },
\end{equation}
where $\overline{{\mathcal L}_{\rm c}}$, $\sigma_{\rm c}$, $\overline{{\mathcal L}_{\rm s}}$, and $\sigma_{\rm s}$ are the mean and the standard deviation of content loss and style loss over $K$ stylized images, respectively.

The quality of stylized images with respect to the criterion (i) is measured using
\begin{equation}\label{eq:length}
    length =\sqrt{ {\widetilde{\mathcal L_{\rm c} } }^{2} + {\widetilde{\mathcal L_{\rm s} } }^{2} }.
\end{equation} 

Let $\omega$  $(\in [0,\frac{\pi}{4}])$ denote the angle between the line going through the origin and $(\widetilde{\mathcal L_{\rm c}}, \widetilde{\mathcal L_{\rm s}})$ and the content loss axis or the style loss axis (the smaller angle is selected):
\begin{equation}\label{eq:angle}
    \omega = \begin{cases}
     \tan^{-1}{\frac{ \widetilde{\mathcal L_{\rm s}} }{ \widetilde{\mathcal L_{\rm c}} }} & {\rm if\ }\widetilde{\mathcal L}_{\rm c} \geq \widetilde{\mathcal L}_{\rm s}\\ 
     \pi/2-\tan^{-1}{\frac{ \widetilde{\mathcal L}_{\rm s} }{ \widetilde{\mathcal L}_{\rm c} }} & {\rm otherwise} 
    \end{cases}.
\end{equation}
Larger $\omega$ indicates that $(\widetilde{\mathcal L_{\rm c}}, \widetilde{\mathcal L_{\rm s}})$ is closer to the balanced axis, meaning that the stylized image is more balanced in content and style. This reflects the criterion (ii).

Using $length$ and $\omega$ above, we define our metric $balance$:
\begin{equation}\label{eq:balance_score}
    balance = \frac{\tan(\omega)}{length}.
\end{equation}
$balance$ concerns both the two criteria (i) and (ii).  
Therefore it is a useful metric for evaluating stylized images. 
We note that larger $balance$ is better because $\tan(\omega)$ should be larger and $length$ should be smaller for better stylized images.

\section{Experimental results} \label{experimental results}
\subsection{Qualitative evaluation} \label{qualitative evaluation}
\begin{figure*}[p]
	\centering
	\includegraphics[width=0.75\linewidth]{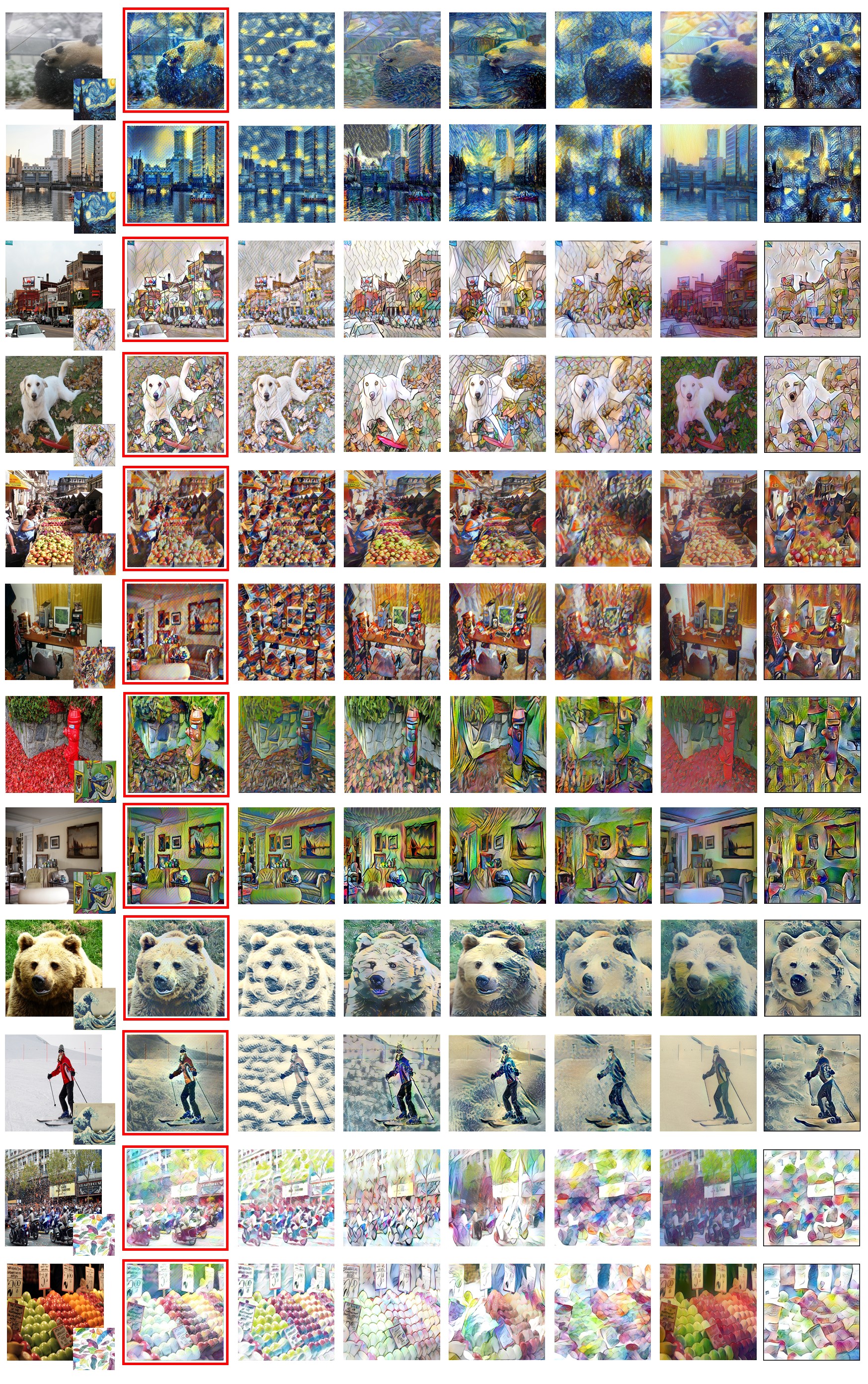}
    \vspace*{-0.5\baselineskip}
	\caption{Visual comparison of our method against the state-of-the-art methods. Left-most column: content image (large) and style image (small). From left to right: the stylized image by our method, Johnson+ \cite{Johnson2016Perceptual}, Huang+ \cite{huang2017adain}, and Gatys+ \cite{gatys2016image}, Sheng+ \cite{sheng2018avatar}, Chen+ \cite{Chen}, and Li+ \cite{Li2017Universal}. Our results surrounded with red rectangles are more balanced in content and style than the others. Note that all stylized images are with the size of $512 \times 512$.} 
	\label{fig:visualization}
\end{figure*}

Figure~\ref{fig:visualization} shows examples of the obtained results, showing that the stylized images obtained by our method are more balanced in content and style. 
We also see that overall the results obtained by Gatys+ \cite{gatys2016image}, Sheng+ \cite{sheng2018avatar}, and Li+ \cite{Li2017Universal} reflect the style well, but they mostly lose content (we cannot understand the content of stylized results using La Muse and Feathers styles). 
In some styles (Starry Night, Composition VII, and The Wave), we see that Johnson+ \cite{Johnson2016Perceptual} seems to randomly select a patch in the style and paste it into the content image. 
Huang+ \cite{huang2017adain} also loses the content and suffers from a so-called checkerboard effect.
We also see that Chen+ \cite{Chen} loses almost style and tends to keep the original content images.

To objectively compare the obtained results, we conducted three user studies, including overall quality, content preserving and style look-like.
From the visual comparison in Fig.~\ref{fig:visualization}, we see that evaluating all stylized results among compared methods is pretty difficult. We thus picked up three methods only for our user studies. 
To this end, we investigated the quantitative comparison (Section~\ref{quantitative evaluation}).
 As Gatys+ \cite{gatys2016image} is known to keep styles most while Johnson+ \cite{Johnson2016Perceptual} retain the content most, these methods are appropriate to choose for our user studies. Among the remaining compared methods, we see that Huang+ \cite{huang2017adain} is most balanced (the loss distributions of Huang+ \cite{huang2017adain} appear near balanced axis (Fig.~\ref{fig:loss_distribution})). We, therefore, chose Gatys+ \cite{gatys2016image}, Johnson+ \cite{Johnson2016Perceptual} and Huang+ \cite{huang2017adain} for our user studies.

For our user studies, we randomly selected 20 images from the 50 testing images as content images and chose 5 styles by excluding The Wave style because it is simpler than the other styles (Fig.~\ref{fig:styles_in_experiment}). We remark that the combination of 20 content images and 5 styles results in 100 stylized images by each method.
In each user study, we presented 100 sets of images to 31 subjects where each set consists of a content image, a style image, and four output images obtained by our method and the three comparison methods~\cite{gatys2016image,huang2017adain,Johnson2016Perceptual}. 
We then asked the subjects to rank the four output stylized images at each set (1st is best, and 4th is worst). 
For the overall quality study, the subjects were asked to give the ranking based on the overall quality at each set. 
For the content preserving study, the subjects were asked to rank output images in each set based on how faithfully the images preserve the content in content images.
For the style look-like study, on the other hand, the subjects ranked output images in each set based on how the images look like the style in style images. 
We note that four output images are aligned in the random order in each set and that each set was displayed for 6 seconds.

Table~\ref{tab:us_quality}, Table~\ref{tab:us_content}, and Table~\ref{tab:us_style} show the average of rankings over the 100 sets for the overall quality, the content preserving, and the style look-like studies, respectively. We also computed the average of rankings in each style, which is also illustrated in Table~\ref{tab:us_quality}, Table~\ref{tab:us_content}, and Table~\ref{tab:us_style}.

We see that our method takes the best ranking among the four methods in overall quality (Table~\ref{tab:us_quality}). Looking into the results in more detail, we see that our method is ranked in the first place at the Mosaic style, and in the second place at others (except for Composition VII style). 
This indicates that our method performs stably well in overall quality in accordance with human cognition.
We remark that the Composition VII style is rather complex (Fig.~\ref{fig:styles_in_experiment}) and, the results for this style are difficult to evaluate.
We also remark that the single-style models (ours, Gatys+\cite{gatys2016image}, and Johnson+\cite{Johnson2016Perceptual}) performed better than the multi-style model (Huang+\cite{huang2017adain}).

For the content preserving (Table~\ref{tab:us_content}) and the style look-like (Table~\ref{tab:us_style}) studies, our method takes the second best ranking.  Note that the scores in these studies more largely distributed than those in the overall quality study.
As MOB-NST is known to perform better in content preserving than IOB-NST \cite{JingYFYS17}; Johnson+ \cite{Johnson2016Perceptual}, which is MOB-NST, takes the best ranking in the content preserving study. 
Gatys+ \cite{gatys2016image}, on the other hand, which is IOB-NST, takes the best ranking in the style look-like study. 
In contrast, our method is ranked in the second place for all styles in the content preserving study (except for Composition VII style) (Table~\ref{tab:us_content}) and in the style look-like study (except for Feathers style) (Table~\ref{tab:us_style}). 
These indicate that our method stably produces stylized images balanced in content and style for almost all the styles.
We remark that in the case of the Feathers style, the two best methods for the look-like study follow the MOB-NST approach.
As MOB-NST is known not to keep styles well \cite{JingYFYS17}, this suggests that the Feathers style is a difficult style for users to evaluate stylized images.

\begin{table}[tb]
\centering
\caption{Average of rankings in the overall quality study. The best and the second best results are given in \textcolor[rgb]{1,0,0}{\textbf{red}} and \textcolor[rgb]{0,0,1}{\textbf{blue}}, respectively.}
\label{tab:us_quality} 
\vspace*{-0.5\baselineskip}
\resizebox{1\linewidth}{!}{%
\begin{tabular}{l@{\hspace*{0.2em}}|@{\hspace*{0.45em}}c@{\hspace*{0.35em}}c@{\hspace*{0.35em}}c@{\hspace*{0.35em}}c}
\toprule\midrule[0.3pt]
\textbf{\centering Style} & \textbf{Ours} & \textbf{Johnson+} & \textbf{Huang+}  & \textbf{Gatys+}\\ 
& & \cite{Johnson2016Perceptual} & \cite{huang2017adain} & \cite{gatys2016image}\\\midrule
Starry Night & \bf \textcolor[rgb]{0,0,1}{2.12} & 2.72 & 3.14 & {\bf \textcolor[rgb]{1,0,0}{2.01}} \\
Mosaic & \bf \textcolor[rgb]{1,0,0}{2.21}  & {\bf \textcolor[rgb]{0,0,1}{2.25}} & 2.91 & 2.63 \\
Composition VII & 2.47 & 2.95 & \bf \textcolor[rgb]{0,0,1}{2.4} & {\bf \textcolor[rgb]{1,0,0}{2.18}} \\
La Muse &  \bf \textcolor[rgb]{0,0,1}{2.38} & {\bf \textcolor[rgb]{1,0,0}{2.28}} & 2.82 & 2.51 \\
Feathers & \bf \textcolor[rgb]{0,0,1}{2.15}  & {\bf \textcolor[rgb]{1,0,0}{1.82}} & 3.28 & 2.74 \\ 
\midrule
All together & \bf \textcolor[rgb]{1,0,0}{2.27} & {\bf \textcolor[rgb]{0,0,1}{2.40}} & 2.91 & 2.41 \\ 
\midrule[0.3pt]\bottomrule
\end{tabular}
}\vspace*{-0.5\baselineskip}
\end{table}

\begin{table}[tb]
\centering
\caption{Average of rankings in the content preserving study. The best and the second best results are given in \textcolor[rgb]{1,0,0}{\textbf{red}} and \textcolor[rgb]{0,0,1}{\textbf{blue}}, respectively.}
\label{tab:us_content} 
\vspace*{-0.5\baselineskip}
\resizebox{1\linewidth}{!}{%
\begin{tabular}{l@{\hspace*{0.2em}}|@{\hspace*{0.45em}}c@{\hspace*{0.35em}}c@{\hspace*{0.35em}}c@{\hspace*{0.35em}}c}
\toprule\midrule[0.3pt]
\textbf{\centering Style} & \textbf{Ours} & \textbf{Johnson+} & \textbf{Huang+}  & \textbf{Gatys+}\\ 
& & \cite{Johnson2016Perceptual} & \cite{huang2017adain} & \cite{gatys2016image}\\\midrule
Starry Night & \bf \textcolor[rgb]{0,0,1}{2.53} & {\bf \textcolor[rgb]{1,0,0}{1.96}} & 2.67 & 2.84 \\
Mosaic & \bf \textcolor[rgb]{0,0,1}{2.13}  & {\bf \textcolor[rgb]{1,0,0}{1.60}} & 3.05 & 3.22 \\
Composition VII & 3.02 & {\bf \textcolor[rgb]{1,0,0}{1.81}} & \bf \textcolor[rgb]{0,0,1}{2.50} & 2.67 \\
La Muse &  \bf \textcolor[rgb]{0,0,1}{1.99} & {\bf \textcolor[rgb]{1,0,0}{1.82}} & 3.06 & 3.13 \\
Feathers & \bf \textcolor[rgb]{0,0,1}{2.02} & {\bf \textcolor[rgb]{1,0,0}{1.81}} & 2.50 & 2.67 \\ 
\midrule
All together & \bf \textcolor[rgb]{0,0,1}{2.34} & {\bf \textcolor[rgb]{1,0,0}{1.80}} & 2.87 & 2.99 \\ 
\midrule[0.3pt]\bottomrule
\end{tabular}
}\vspace*{-0.5\baselineskip}
\end{table}

\begin{table}[tb]
\centering
\caption{Average of rankings in the style look-like study. The best and the second best results are given in \textcolor[rgb]{1,0,0}{\textbf{red}} and \textcolor[rgb]{0,0,1}{\textbf{blue}}, respectively.}
\label{tab:us_style} 
\vspace*{-0.5\baselineskip}
\resizebox{1\linewidth}{!}{%
\begin{tabular}{l@{\hspace*{0.2em}}|@{\hspace*{0.45em}}c@{\hspace*{0.35em}}c@{\hspace*{0.35em}}c@{\hspace*{0.35em}}c}
\toprule\midrule[0.3pt]
\textbf{\centering Style} & \textbf{Ours} & \textbf{Johnson+} & \textbf{Huang+}  & \textbf{Gatys+}\\ 
& & \cite{Johnson2016Perceptual} & \cite{huang2017adain} & \cite{gatys2016image}\\\midrule
Starry Night & \bf \textcolor[rgb]{0,0,1}{2.27} & 2.77 & 3.34 &  {\bf \textcolor[rgb]{1,0,0}{1.61}} \\
Mosaic & \bf \textcolor[rgb]{0,0,1}{2.26} & 2.66 & 2.94 &  {\bf \textcolor[rgb]{1,0,0}{2.13}} \\
Composition VII & \bf \textcolor[rgb]{0,0,1}{2.49} & 2.96 & 2.65 &  {\bf \textcolor[rgb]{1,0,0}{1.90}} \\
La Muse & \bf \textcolor[rgb]{0,0,1}{2.71} & 2.81 & 2.81 &  {\bf \textcolor[rgb]{1,0,0}{1.67}} \\
Feathers & \bf \textcolor[rgb]{1,0,0}{1.69} & {\bf \textcolor[rgb]{0,0,1}{2.34}}& 3.42 & 2.55 \\ 
\midrule
All together & \bf \textcolor[rgb]{0,0,1}{2.28} & 2.71 & 3.03 &  {\bf \textcolor[rgb]{1,0,0}{1.97}} \\ 
\midrule[0.3pt]\bottomrule
\end{tabular}
}\vspace*{-0.5\baselineskip}
\end{table}

\subsection{Quantitative evaluation} \label{quantitative evaluation}


\begin{table}[tb]
\centering
\caption{Averages of $length$ (smaller is better) and $balance$ (larger is better).} 
\label{tab:length_balance_comparison}
\vspace*{-0.5\baselineskip}
\resizebox{0.9\linewidth}{!}{
\begin{tabular}{l@{\hspace*{0.2em}}|c@{\hspace*{0.35em}}c@{\hspace*{0.35em}}c@{\hspace*{0.35em}}c@{\hspace*{0.35em}}}
\toprule[1pt]\midrule[0.3pt]
\textbf{Method} & $length$ ($\Downarrow$) & \quad & \quad & $balance$ ($\Uparrow$) \\ \midrule
\textbf{Ours} & \textbf{0.37} & \quad & \quad & \textbf{2.95} \\
\textbf{Johnson+~\cite{Johnson2016Perceptual}} \quad \quad &  0.54 & \quad & \quad & 1.60 \\
\textbf{Huang+~\cite{huang2017adain}} & 0.45 & \quad & \quad & 1.23 \\
\textbf{Gatys+~\cite{gatys2016image}} & 0.45 & \quad & \quad & 1.36 \\
\textbf{Sheng+~\cite{sheng2018avatar}} & 0.52 & \quad & \quad & 1.21 \\
\textbf{Chen+~\cite{Chen}} & 0.59 & \quad & \quad & 0.72\\
\textbf{Li+~\cite{Li2017Universal}} & 0.49 & \quad & \quad & 1.40\\
\midrule[0.3pt]\bottomrule[1pt]
\end{tabular}}
\end{table}

\begin{figure*}[tb]
    \centering
    \begin{subfigure}[b]{0.45\textwidth}
        \includegraphics[width=\textwidth]{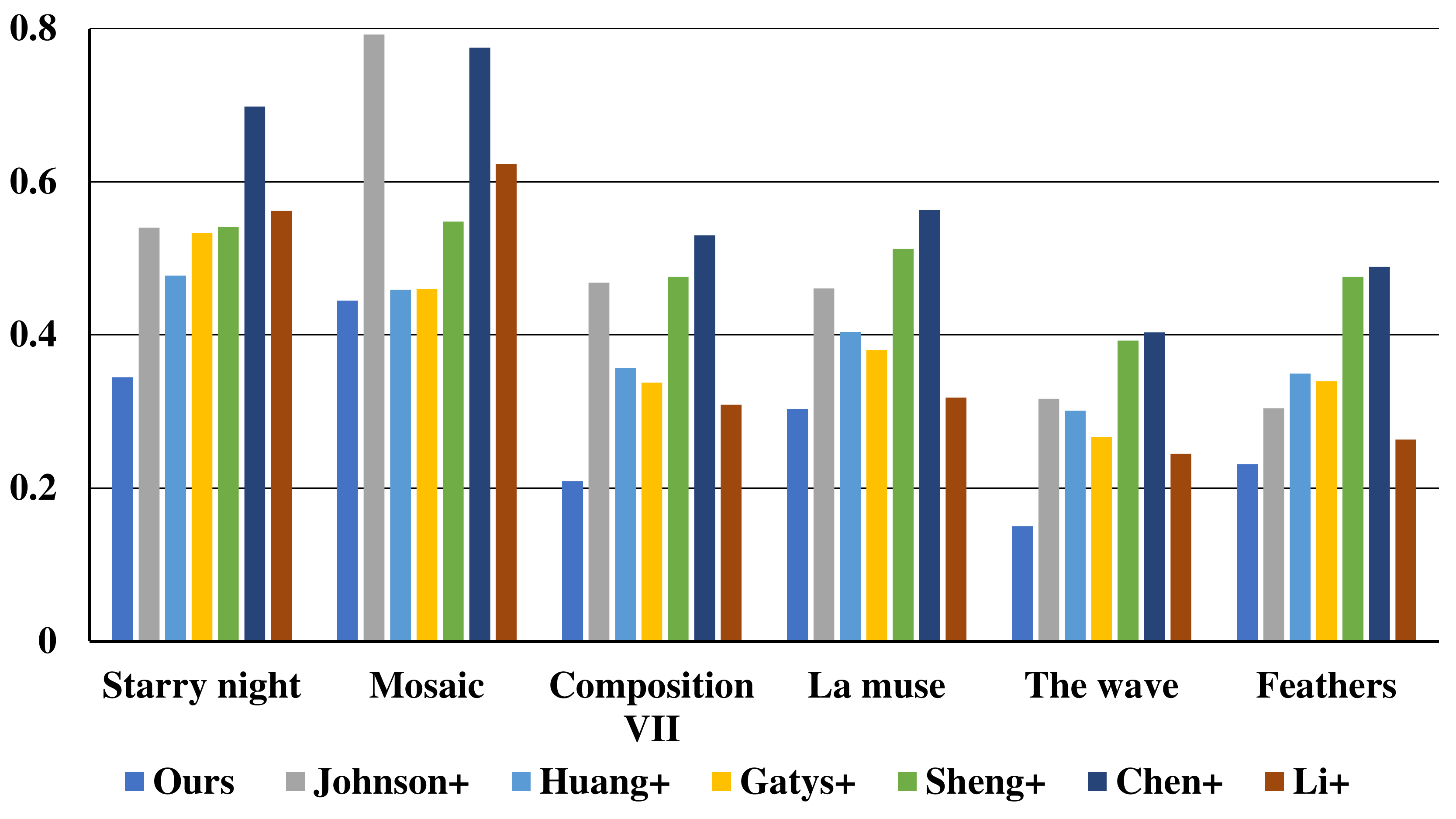}
        \vspace*{-\baselineskip}
        \caption{$length$ ($\Downarrow$).}
        \label{fig:length_in_styles}
    \end{subfigure}
    \begin{subfigure}[b]{0.45\textwidth}
        \includegraphics[width=\textwidth]{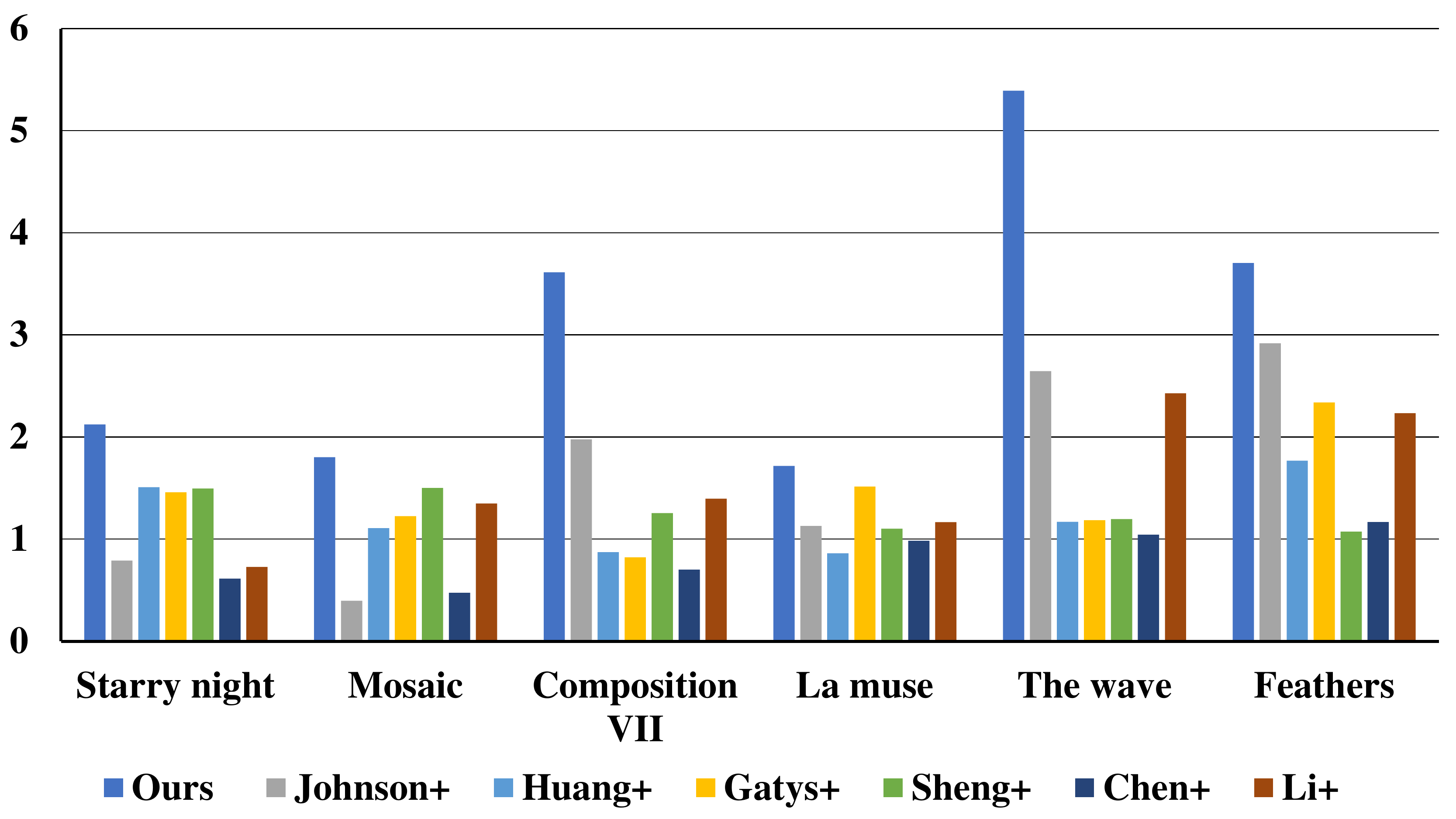}
        \vspace*{-\baselineskip}
        \caption{$balance$ ($\Uparrow$).}
        \label{fig:balance_score_in_styles}
    \end{subfigure}
    \caption{Averages of $length$ and $balance$ in each style.} 
    \label{fig:length_balance_in_styles}
    \vspace*{-0.5\baselineskip}
\end{figure*}

\begin{figure*}[tb]
    \centering
    \begin{subfigure}[b]{0.3\textwidth}
        \includegraphics[width=\textwidth]{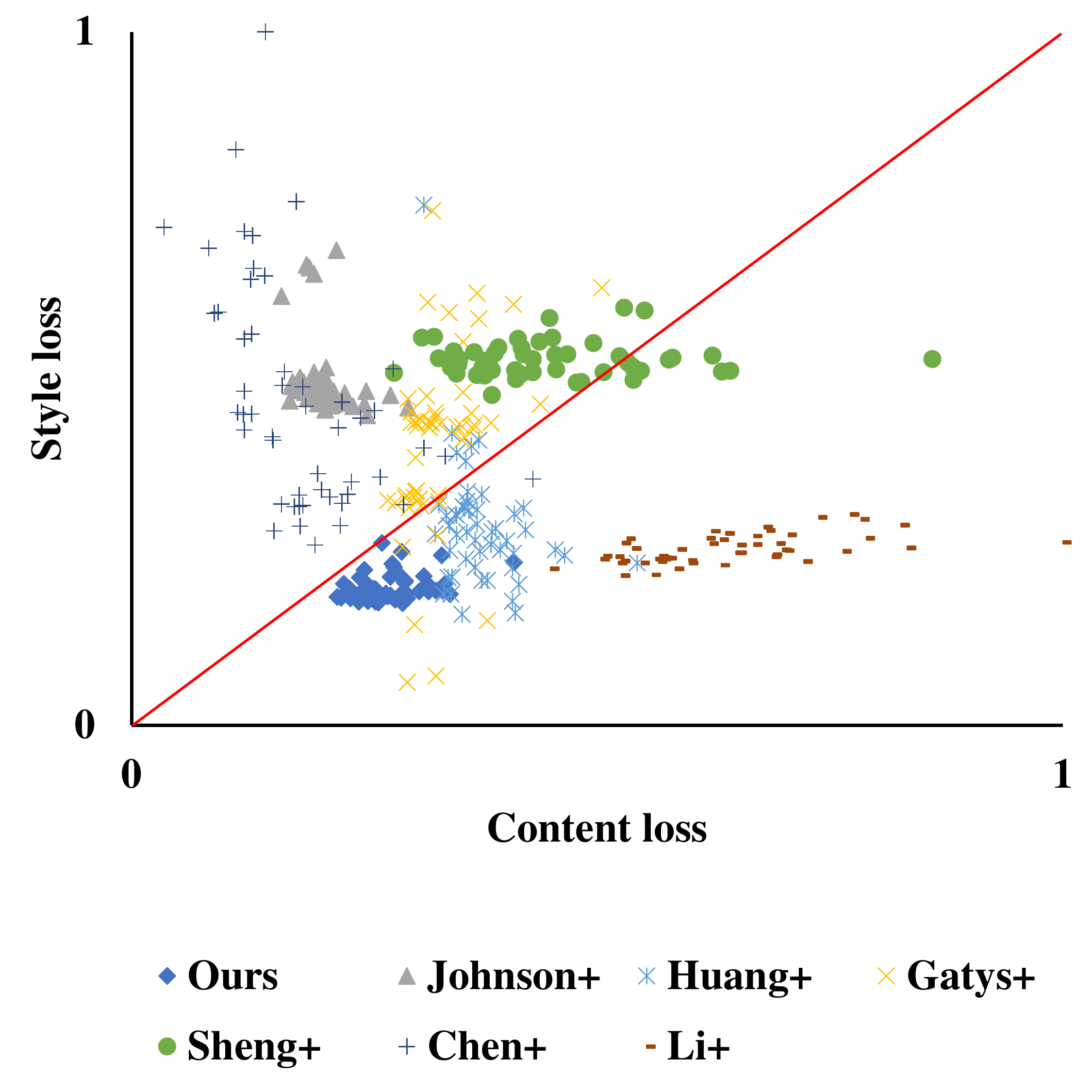}
        \caption{Starry Night style.}
        \label{fig:starry_night_scale}
    \end{subfigure}
    \begin{subfigure}[b]{0.3\textwidth}
        \includegraphics[width=\textwidth]{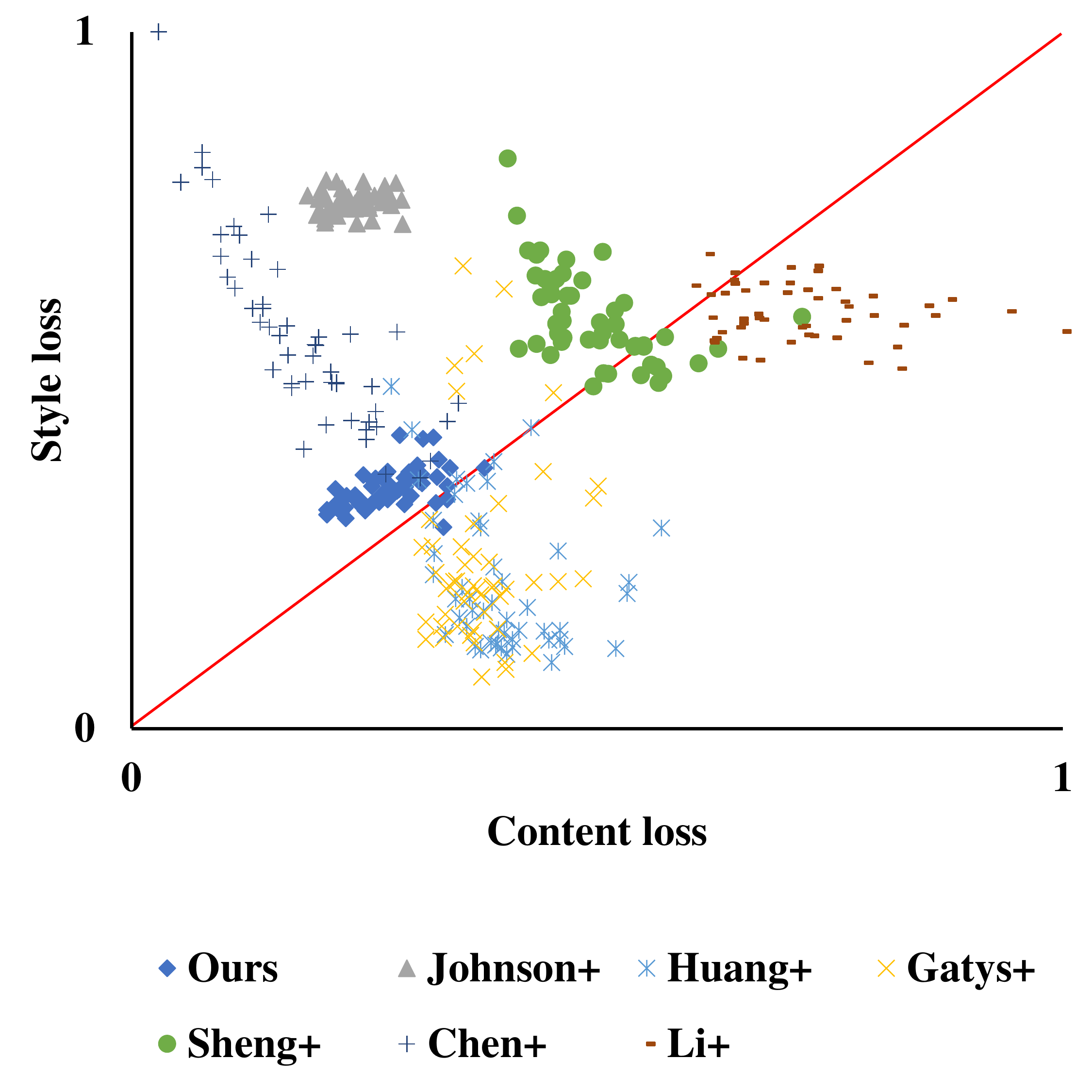}
        \caption{Mosaic style.}
        \label{fig:mosaic_scale}
    \end{subfigure}
    \begin{subfigure}[b]{0.3\textwidth}
        \includegraphics[width=\textwidth]{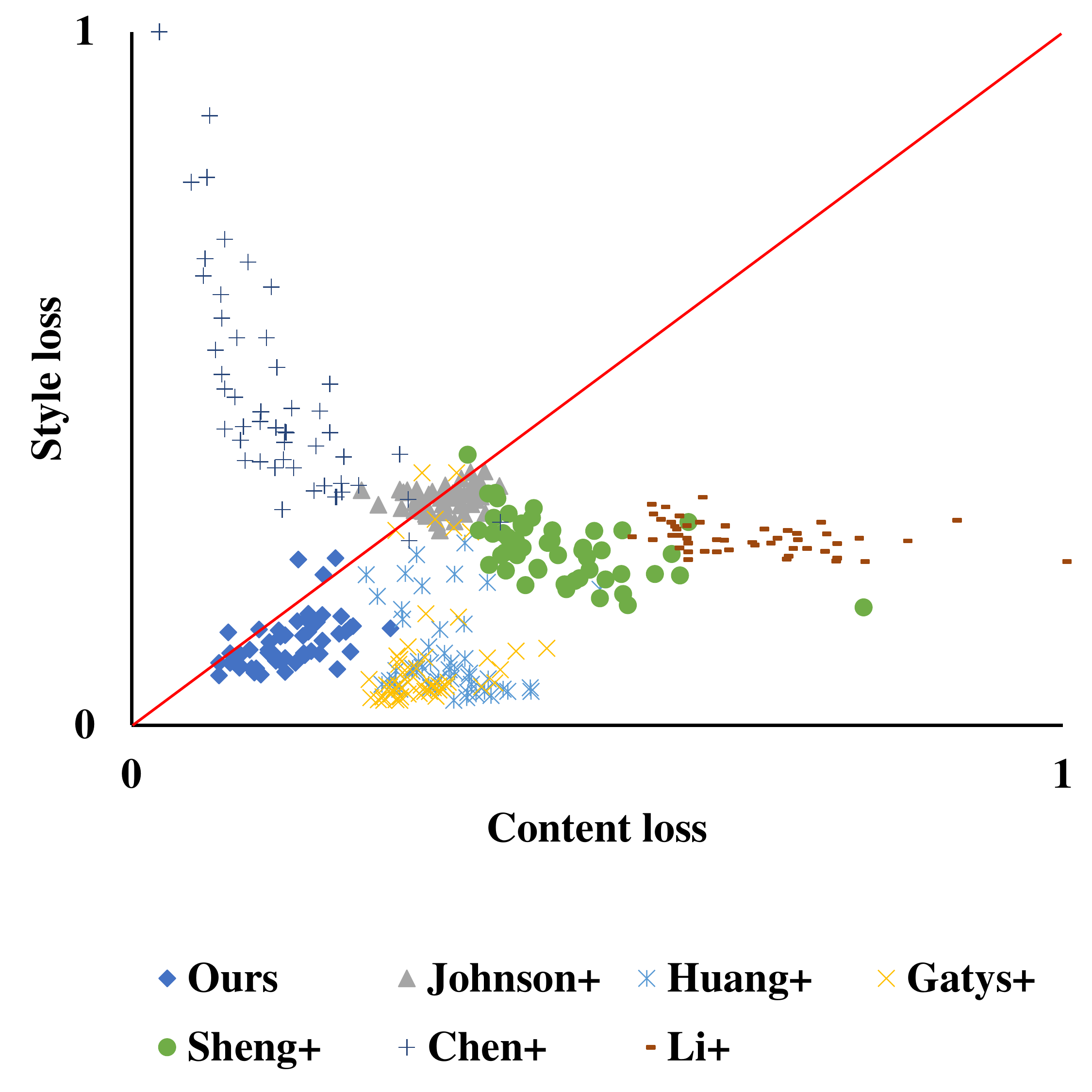}
        \caption{Composition VII style.}
        \label{fig:composition_scale}
    \end{subfigure}
    \begin{subfigure}[b]{0.3\textwidth}
        \includegraphics[width=\textwidth]{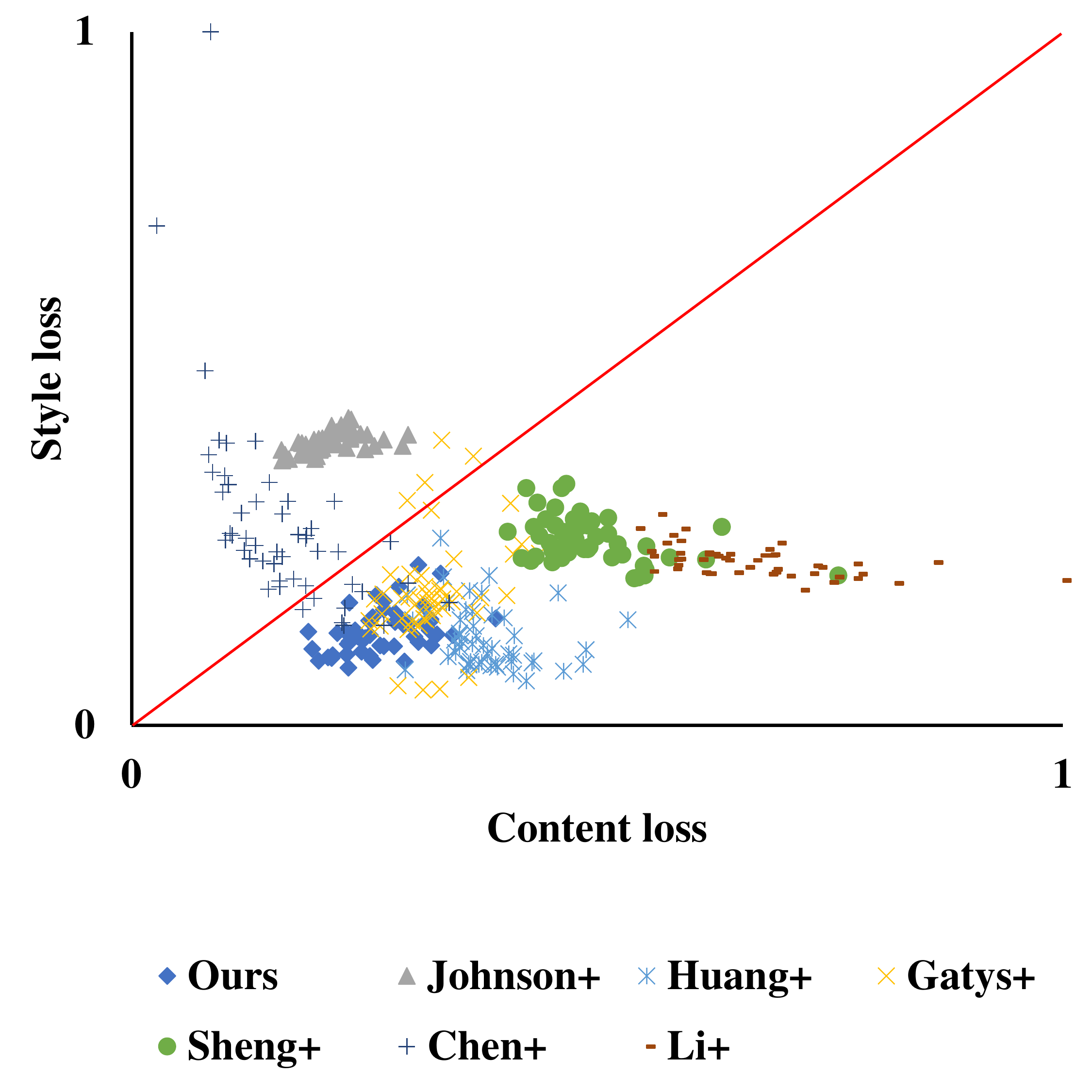}
        \caption{La Muse style.}
        \label{fig:muse_scale}
    \end{subfigure}
    \begin{subfigure}[b]{0.3\textwidth}
        \includegraphics[width=\textwidth]{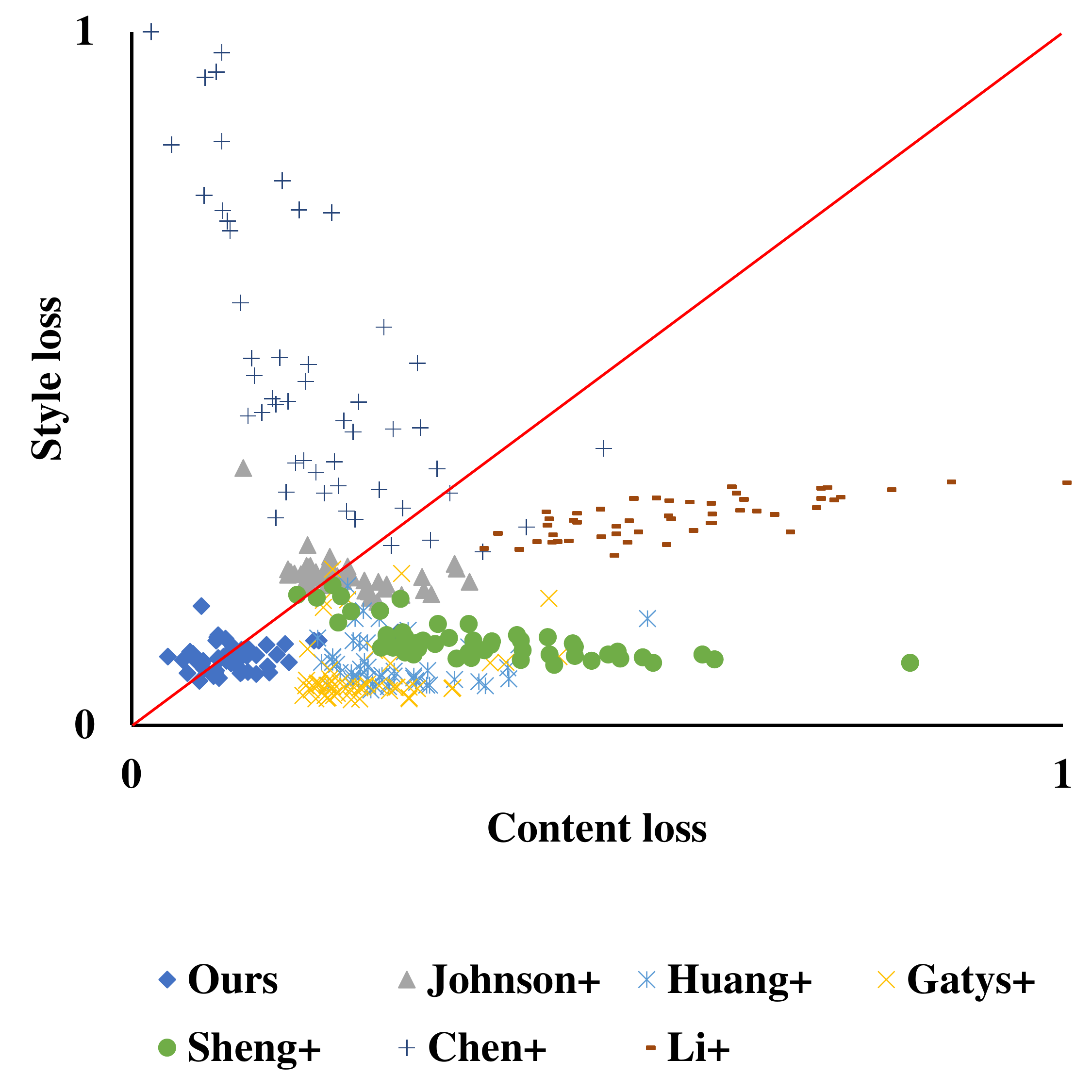}
        \caption{The Wave style.}
        \label{fig:wave_scale}
    \end{subfigure}
    \begin{subfigure}[b]{0.3\textwidth}
        \includegraphics[width=\textwidth]{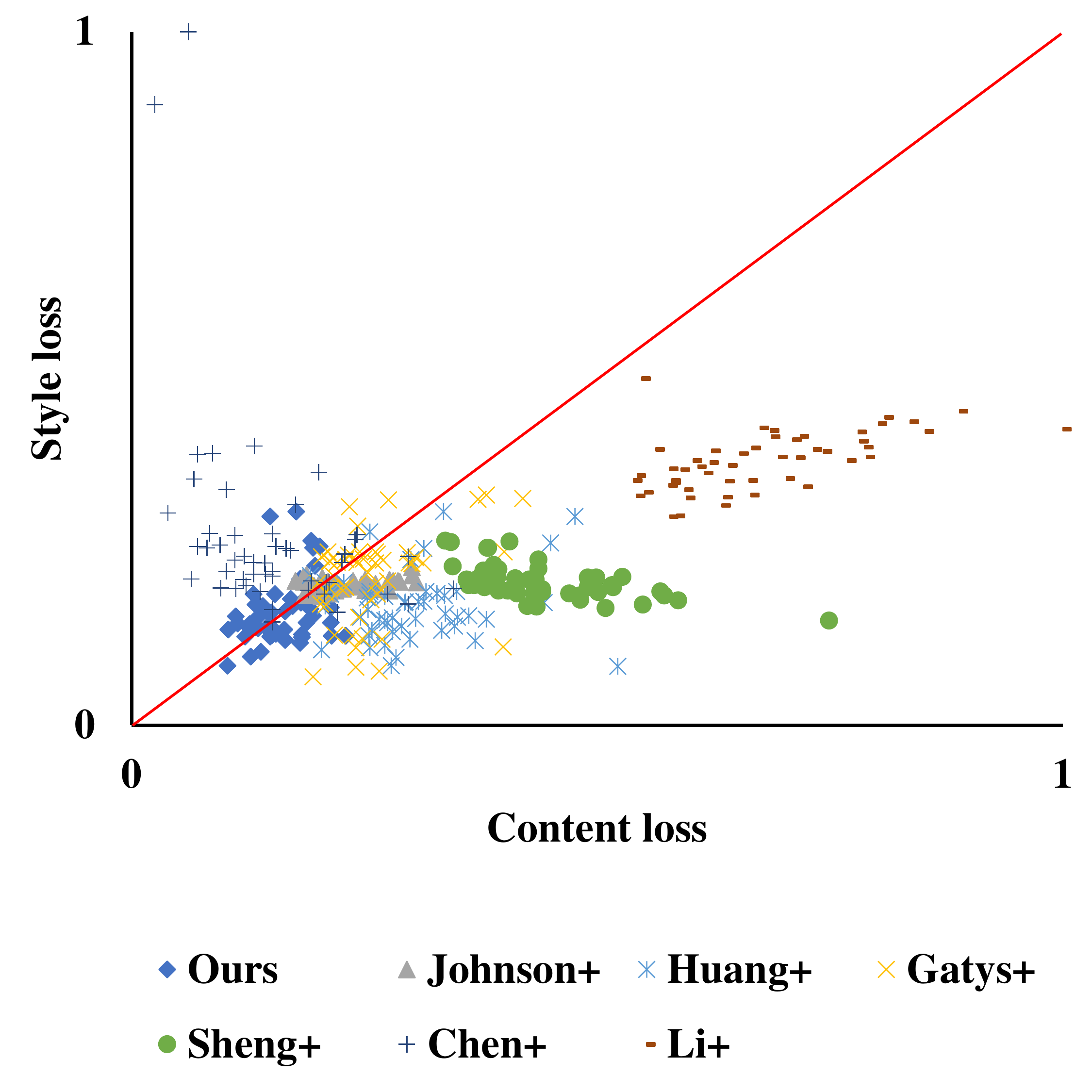}
        \caption{Feathers style.}
        \label{fig:features_scale}
    \end{subfigure}
    \caption{Loss distribution in each style.
    Red lines denote the balanced axis. Our method has the distributions nearer the balanced axis than the other methods.}\label{fig:loss_distribution}
\end{figure*}

In order to quantitatively evaluate the obtained results, we computed
the averages of $length$'s and $balance$'s over 300 ($=50$ contents $\times$ 6 styles) sets for each method (Table~\ref{tab:length_balance_comparison}). 
We see that our method performs best both in $length$ and $balance$.
We also computed the averages of $length$'s and $balance$'s in each style, which is illustrated in
Fig.~\ref{fig:length_balance_in_styles}.
Fig.~\ref{fig:length_balance_in_styles} shows that our method performs best in $length$ and  best in $balance$ for all the styles.

To look into the results in more detail, we show the loss distribution of 50 stylized images in each style (Fig.~\ref{fig:loss_distribution}). We see that (1) the content loss and the style loss (for each stylized result) in our method are similar with each other and that (2) loss distributions in our method appear densely near the balanced axis for all the styles while those in the other methods do not.

\subsection{Computational speed}

\begin{table}[tb]
\centering
\caption{The average wall-clock time in second for producing one stylized image.}
\vspace*{-0.5\baselineskip}
\label{tab:speed}
\resizebox{1\linewidth}{!}{
\begin{tabular}{l@{\hspace*{0.2em}}|c@{\hspace*{0.35em}}c@{\hspace*{0.35em}}c@{\hspace*{0.35em}}c@{\hspace*{0.35em}}c@{\hspace*{0.35em}}}
\toprule[1pt]\midrule[0.3pt]
\multirow{2}{*}{\textbf{Method}} & \multicolumn{4}{c}{\textbf{Image size}} & \textbf{Implemented framework} \\ 
& $256 \times 256$ & \quad & \quad & $512 \times 512$ \\ \midrule
\textbf{Ours} & \textbf{0.05} & \quad & \quad & \textbf{0.18} & PyTorch \\
\textbf{Johnson+~\cite{Johnson2016Perceptual}} \quad \quad & 1.12 & \quad & \quad & 3.79 & Torch\\
\textbf{Huang+~\cite{huang2017adain}} & 1.98 & \quad & \quad & 6.78 & Torch \\
\textbf{Gatys+~\cite{gatys2016image}} & 74.12 & \quad & \quad & 269.74 & Torch\\
\textbf{Sheng+~\cite{sheng2018avatar}} & 3.04 & \quad & \quad & 10.67 & TensorFlow\\
\textbf{Chen+~\cite{Chen}} & 2.74 & \quad & \quad & 9.33 & Torch\\
\textbf{Li+~\cite{Li2017Universal}} & 3.53 & \quad & \quad & 9.42 & Torch\\
\midrule[0.3pt]\bottomrule[1pt]
\end{tabular}
}
\end{table}

We measured the running time for generating 300 stylized images with the sizes of $256 \times 256$ and $512 \times 512$ by each method and compared the average for generating one stylized image by each method.

Table~\ref{tab:speed} illustrates the average of the running time in generating one stylized image.
As we see, our method is the fastest and speeds up 22 times for the image size of $256\times256$ and 21 times for that of $512\times512$ when compared with the fastest state-of-the-arts~\cite{Johnson2016Perceptual}.
We can thus conclude that our method is promising for real-time applications.

\subsection{More detailed analysis}

\subsubsection{Behavior of balance weight $\gamma$ during the training}

\begin{figure}[tb]
	\centering
	\includegraphics[width=1\linewidth]{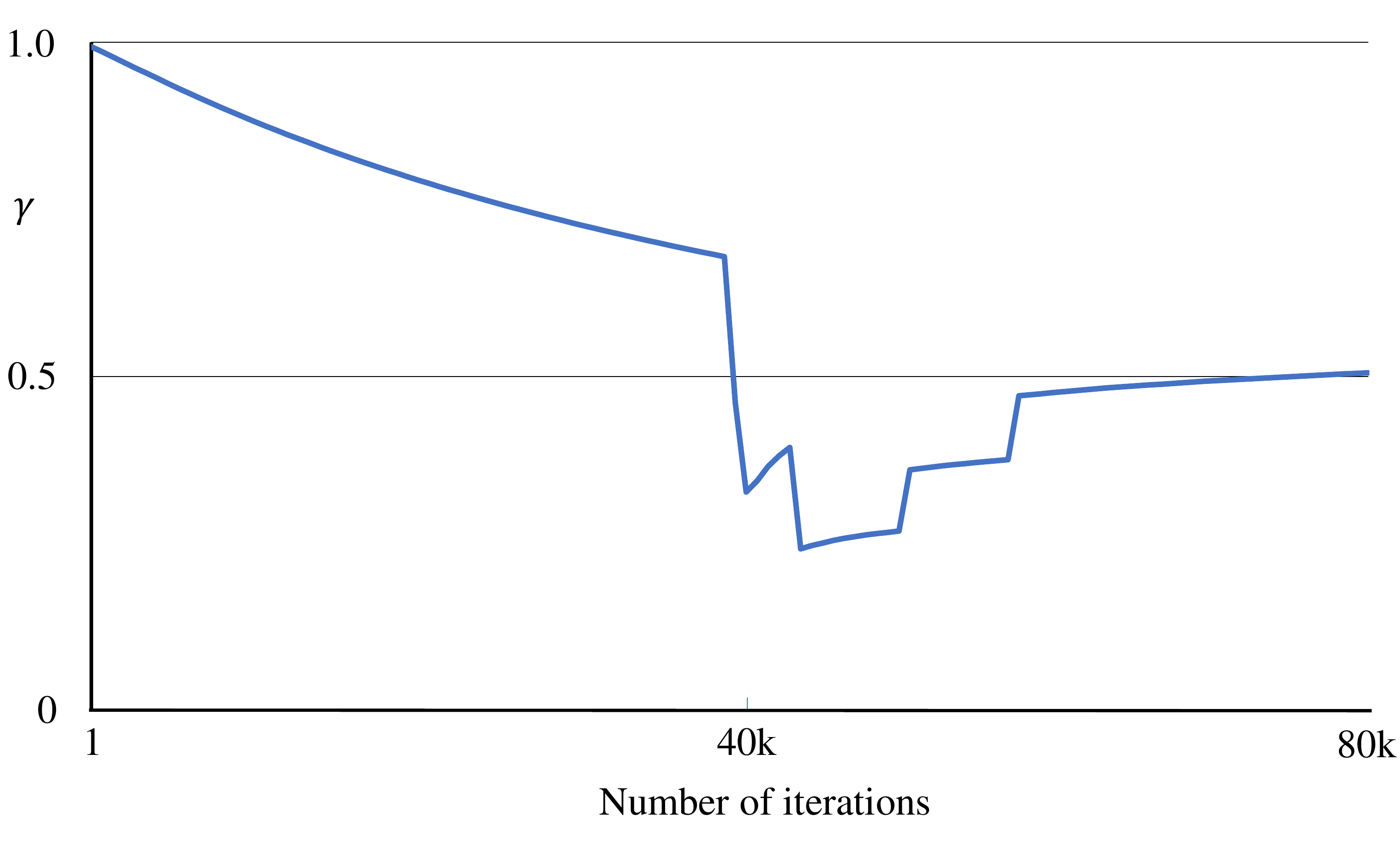}
	\caption{Behavior of $\gamma$ during the training on the Starry Night style.} 
	\label{fig:gamma}
\end{figure}


We investigate the behavior of balance weight $\gamma$ to verify that $\gamma$ is adaptively updated to converge to an expected value.

Figure~\ref{fig:gamma} illustrates how balance weight $\gamma$ changes during the training on the Starry Night style.
We see that $\gamma$ is adaptively updated corresponding to the content and style losses.
We remark that since we set $\alpha=0.5$ in the loss function, $\gamma$ is expected to be close to 0.5 after the training.
At the beginning of training, the style loss $\mathcal{L}_s$ is far larger than the content loss $\mathcal{L}_c$, resulting in $\gamma$ far larger than $0.5$ (close to 1.0).
As the training proceeds, the network is gradually optimized, resulting $\gamma$ close to 0.5 in the end of the training.

We observe that $\gamma$ quickly decreases after one epoch (about 40k iterations).
This can be explained as follows.
After one epoch, the overfitting problem on the style image occurs since our network is trained using a single style image. Hence, the style loss quickly drops.
As a result, the behavior of $\gamma$ becomes different.
Indeed, we observed that the style loss raised the overfitting problem through the validation phase. We thus stopped the training of the style subnet while kept updating the weights of the other subnets. As a result, the content loss decreased more quickly than the style loss.
Then, $\gamma$ was gradually recovered; its value became close to 0.5 in the end of training.

This evaluation confirms that $\gamma$ gradually adapts to achieve the equal contributions of content and style in stylized images during the training thanks to our adaptive feature injection and concatenation.
We remark that we observed similar behaviors of $\gamma$ for other styles.


\subsubsection{Effectiveness of feature injection}


\begin{figure*}[!t]
	\centering
	\includegraphics[width=\linewidth]{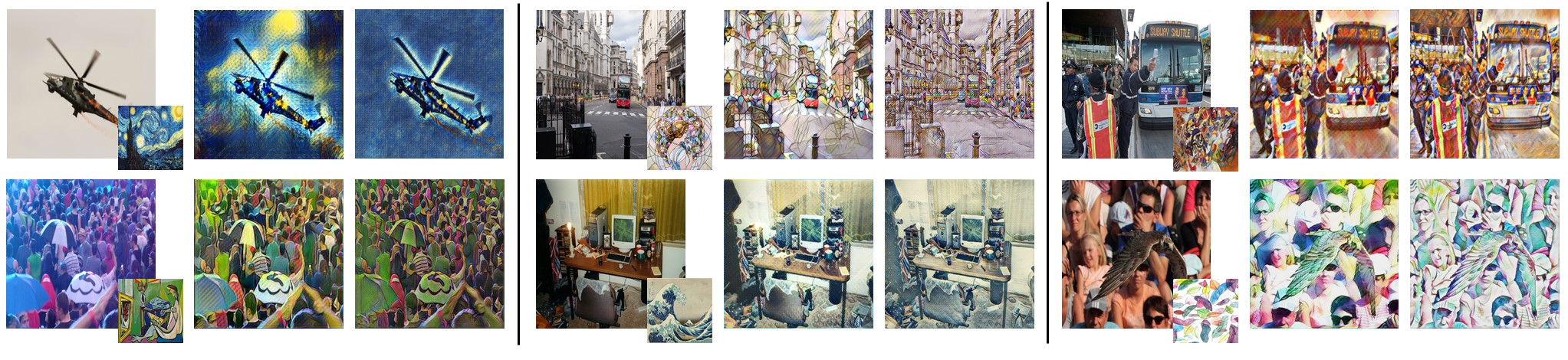}
    \vspace*{-0.5\baselineskip}
	\caption{Visual comparison of the complete model and the model w/o feature injection. In each block, from left to right, a content image (large one) with a style (small one) is followed by outputs by the complete model and the model w/o feature injection. Note that all stylized images are with the size of $512 \times 512$.} 
	\label{fig:visualization_detail}
\end{figure*}

\begin{figure*}[tb]
    \centering
    \begin{subfigure}[b]{0.3\textwidth}
        \includegraphics[width=\textwidth]{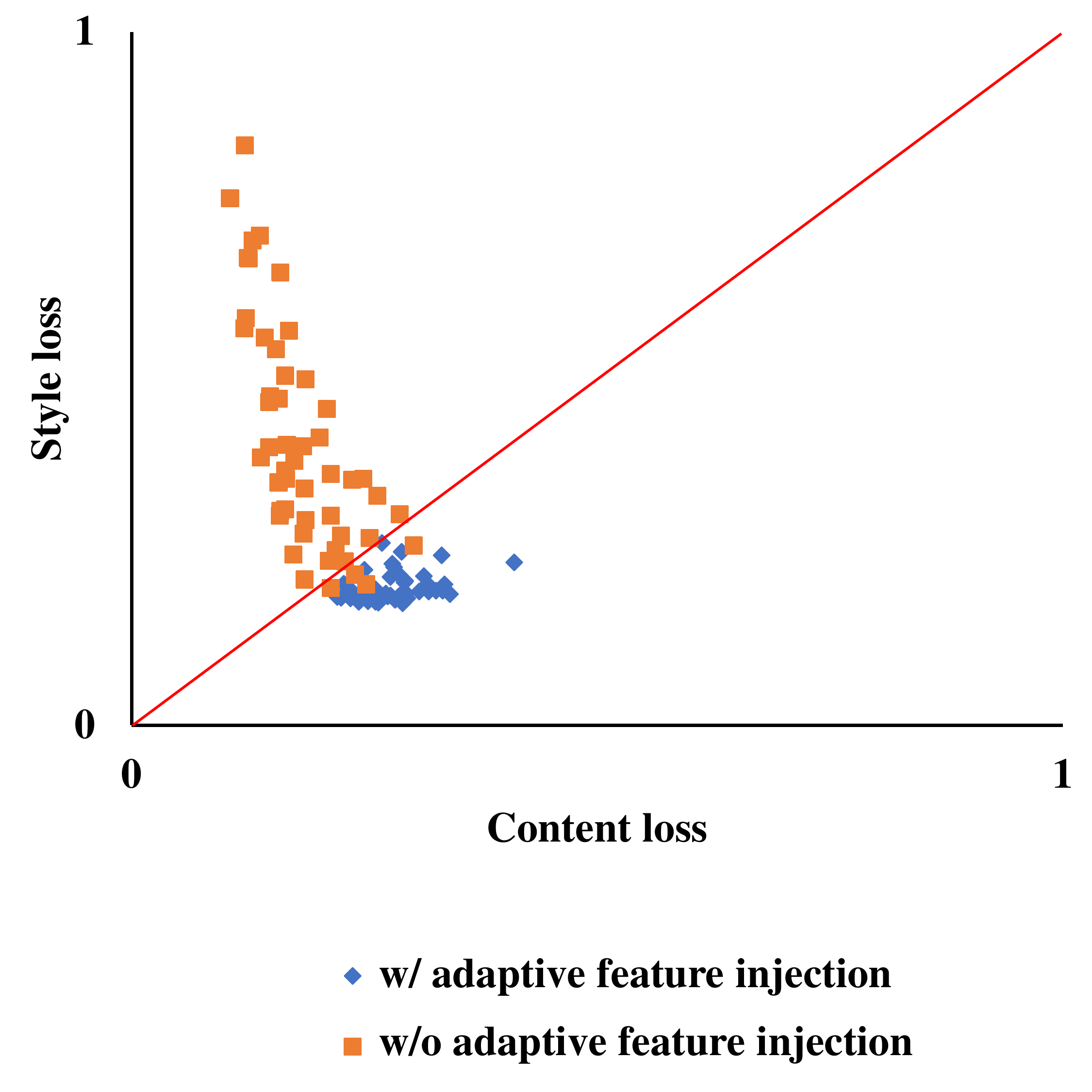}
        \caption{Starry Night style.}
        \label{fig:starry_night_scale_detail}
    \end{subfigure}
   \begin{subfigure}[b]{0.3\textwidth}
        \includegraphics[width=\textwidth]{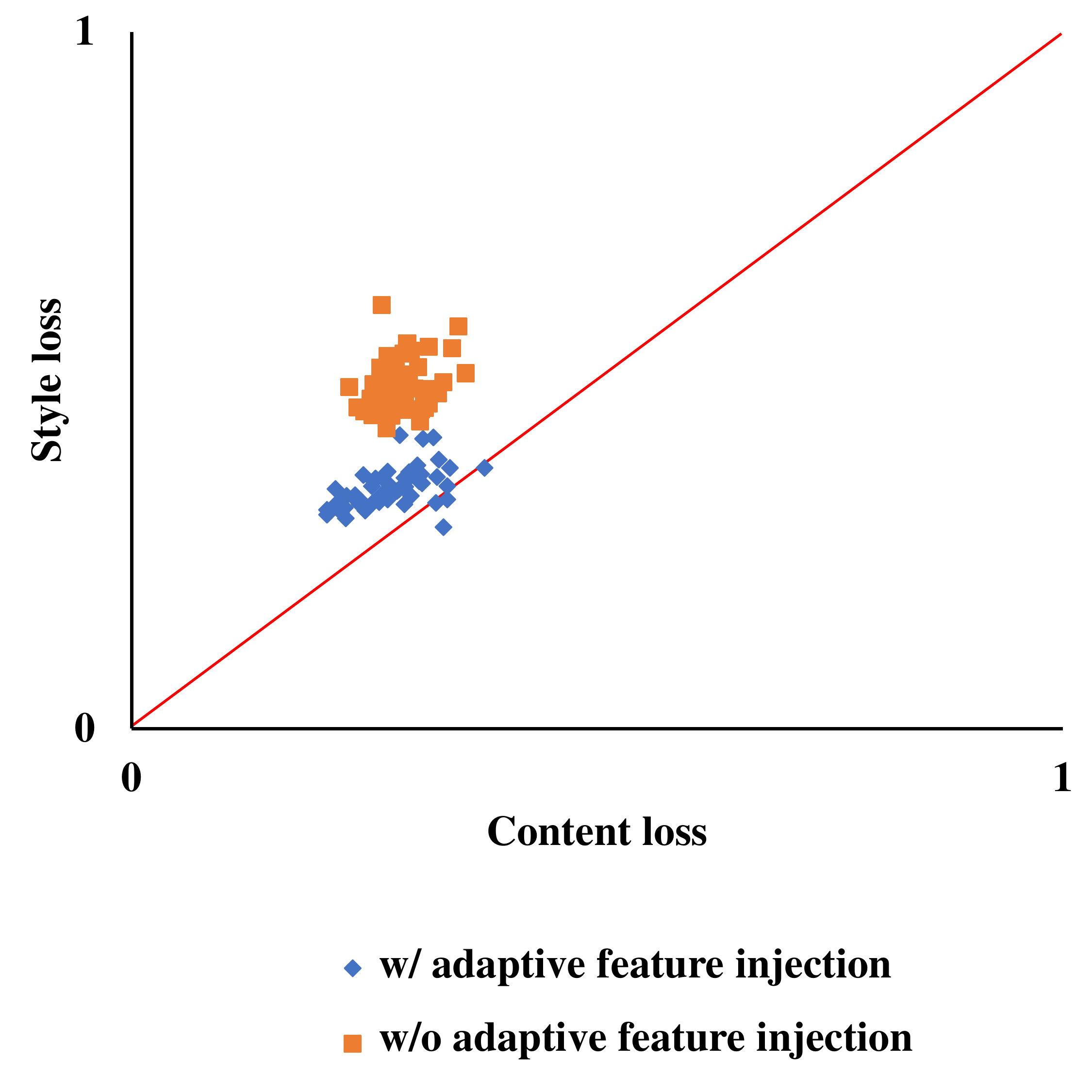}
        \caption{Mosaic style.}
        \label{fig:mosaic_scale_detail}
    \end{subfigure}
    \begin{subfigure}[b]{0.3\textwidth}
        \includegraphics[width=\textwidth]{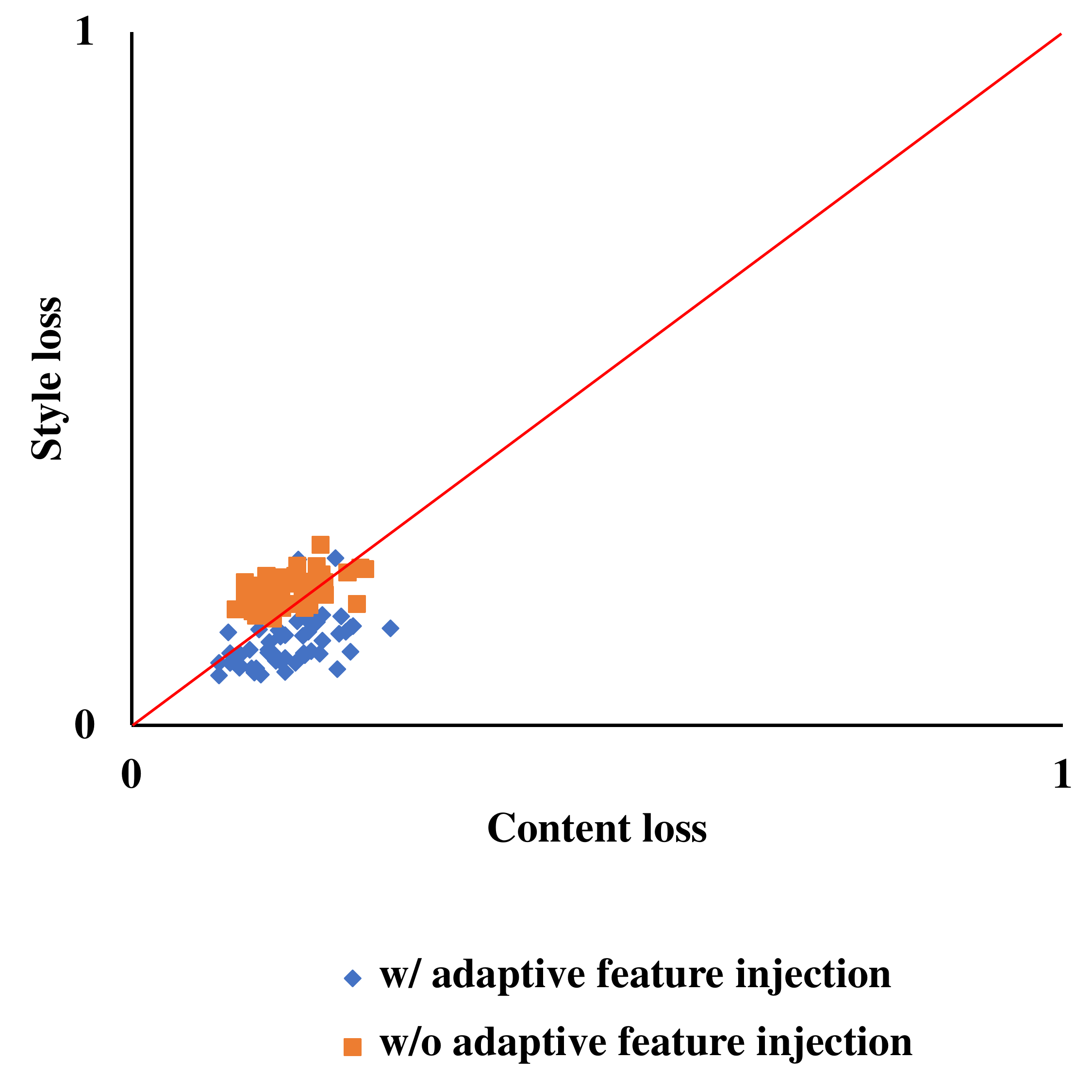}
        \caption{Composition VII style.}
        \label{fig:composition_scale_detail}
    \end{subfigure}
   \begin{subfigure}[b]{0.3\textwidth}
        \includegraphics[width=\textwidth]{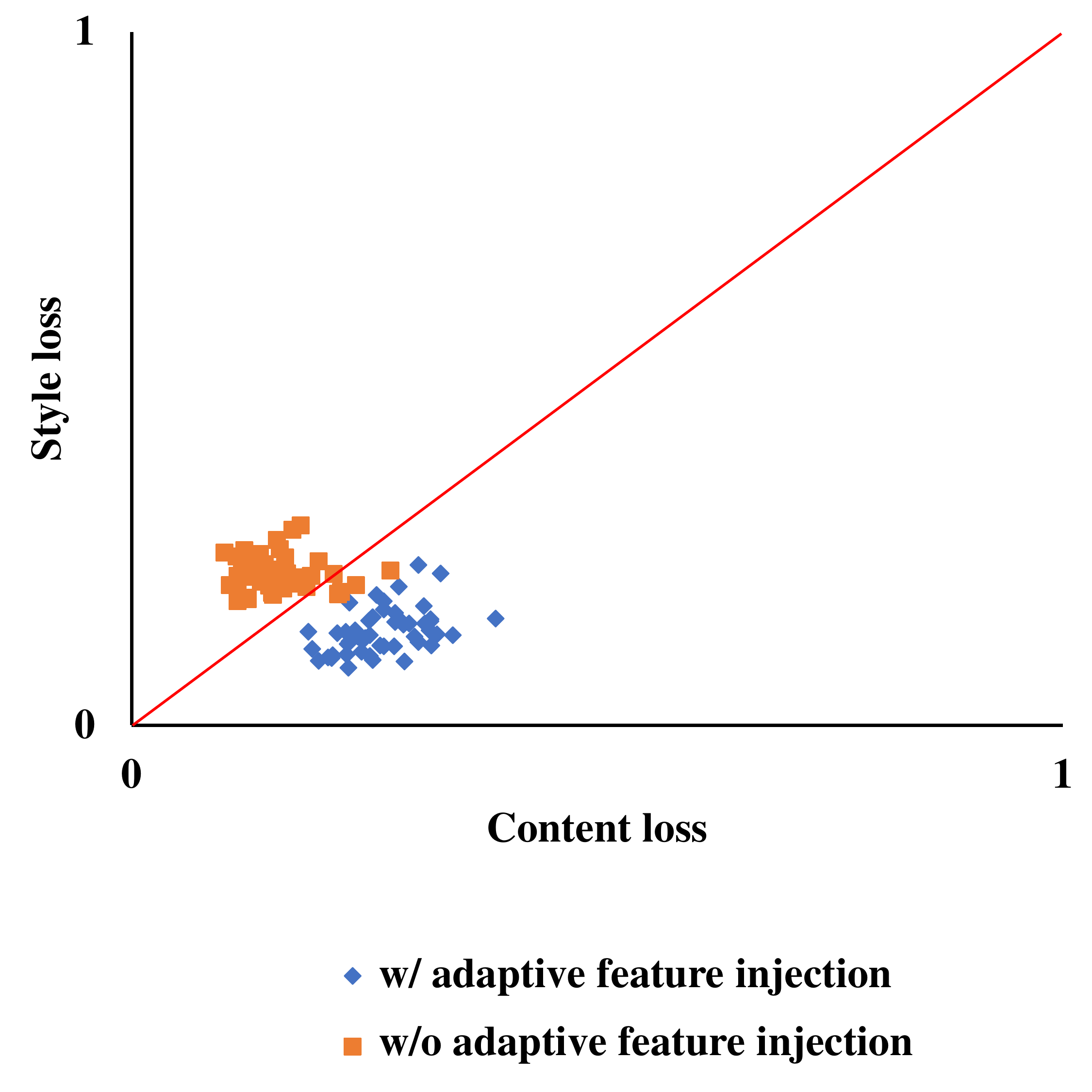}
        \caption{La Muse style.}
        \label{fig:lamuse_scale_detail}
    \end{subfigure}
    \begin{subfigure}[b]{0.3\textwidth}
        \includegraphics[width=\textwidth]{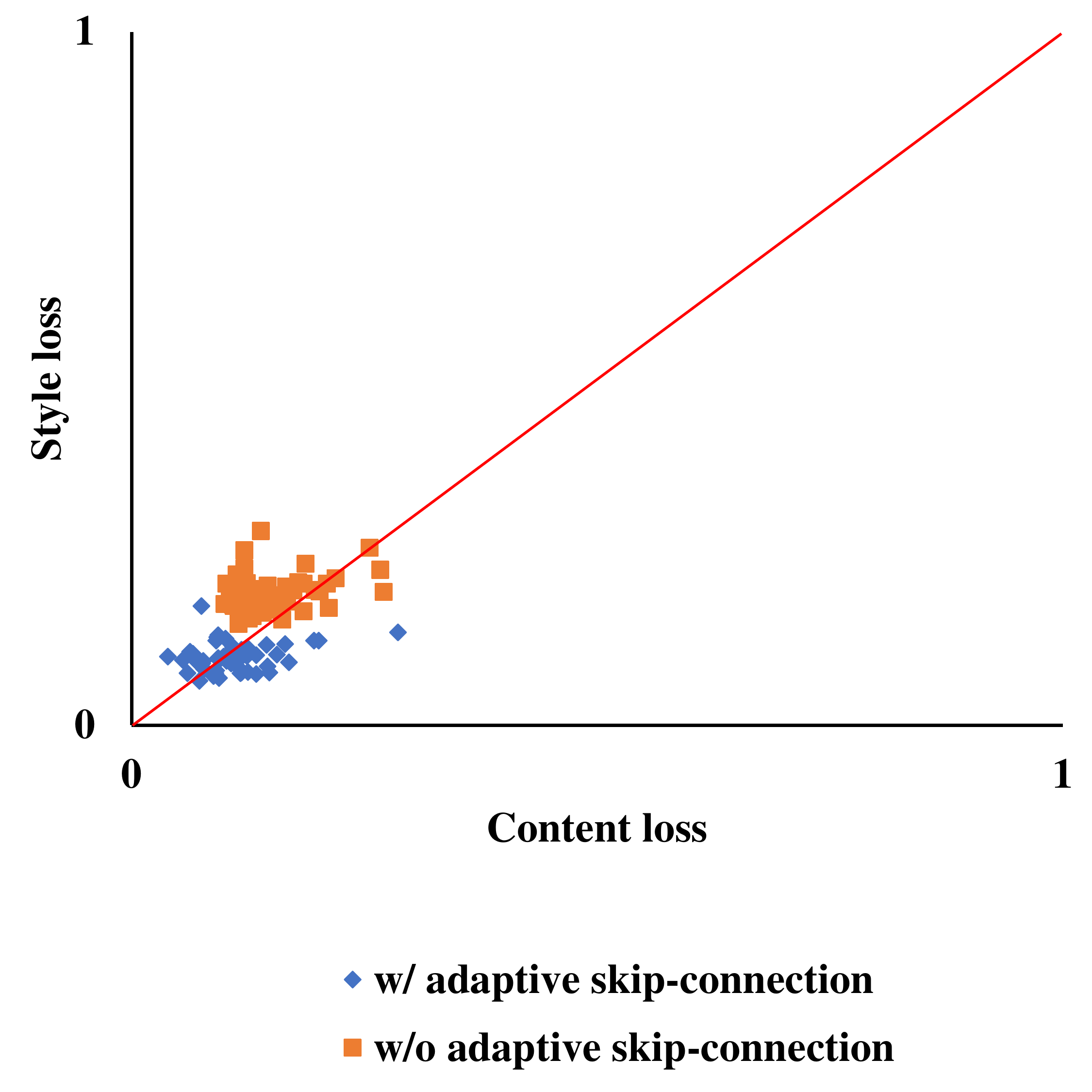}
        \caption{The Wave style.}
        \label{fig:wave_scale_detail}
    \end{subfigure}
     \begin{subfigure}[b]{0.3\textwidth}
        \includegraphics[width=\textwidth]{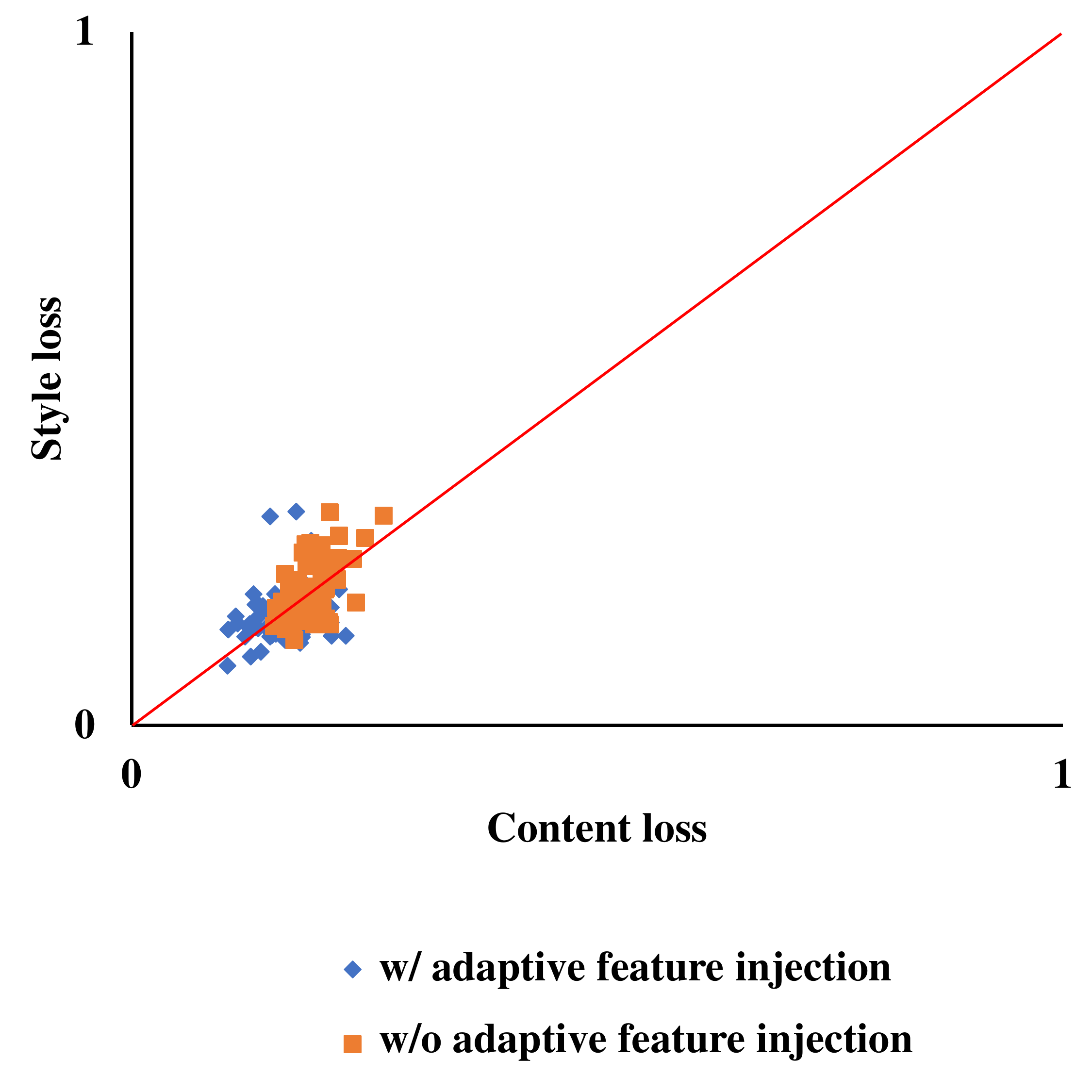}
        \caption{Feathers style.}
        \label{fig:feathers_scale_detail}
    \end{subfigure}
    \caption{Loss distribution in each style obtained by the complete model and the model w/o feature injection.
    Red lines denote the balanced axis.}\label{fig:loss_distribution_detail}
\end{figure*}

In this section, we evaluate the effectiveness of the introduction to the adaptive feature injection between the content subnet and the style subnet.

We compared our complete model with the model w/o feature injection (i.e., the model that disabled only the adaptive feature injection), which is shown in Fig.~\ref{fig:visualization_detail}.
Fig.~\ref{fig:visualization_detail} shows that the stylized images obtained by the complete model are in general more balanced in content and style than those by the model w/o feature injection.
However, we can see roughly global structure appearing in the synthesized images in Fig.~\ref{fig:visualization_detail} upper set (in particular, the leftmost which is with Starry Night style).
This can be explained as follows.
In general, the model w/o feature injection tends to preserve more content than style while 
the complete model does more style than content.
This is because the feature injection from the style subnet to the content subnet tries to reduce the style loss (see below).
The feature injection at multiple layers employed in the content and style subnets helps to keep both global and local structure in rendering.
As a result, global structure in stylized images such as the stroke in the Starry Night may sometimes become impressive.

We also compared the $length$ and $balance$ of stylized images (Table~\ref{tab:length_balance_comparison_detail}). 
We see that the complete model performs better both in $length$ and $balance$ than the model w/o feature injection. 
Table~\ref{tab:length_balance_comparison_detail} also shows that employing adaptive feature injection improves both $length$ and $balance$ for each style (except for La Muse style). 
This indicates that adaptive feature injection is effective to improve not only the quality but also the balance in content and style of stylized images.
With respect to the La Muse style, $length$ of the complete model is comparable to that of model w/o feature injection, however $balance$ is not the case. 
This can be explained as follows.
The La Muse style follows Cubism and thus it is very unique. Because of this, the adaptive feature injection tends to keep more style to reflect the impression of this style.

Finally, we compare the loss distributions of 50 stylized images in each style (Fig.~\ref{fig:loss_distribution_detail}). 
We see that for all styles (except for the La Muse style) the loss distributions of the complete model appears more densely near the balanced axis and is closer to the origin than those of the model w/o feature injection for all styles. 
In the case of the Starry Night style (Fig.~\ref{fig:starry_night_scale_detail}), we see that the model w/o feature injection preserves much more content than the style because the loss distribution appears far above the balanced axis. 
This observation also holds true for the Mosaic style (Fig.~\ref{fig:mosaic_scale_detail}), the Composition VII (Fig.~\ref{fig:composition_scale_detail}), and the La Muse (Fig.~\ref{fig:lamuse_scale_detail}). 
By using adaptive feature injection, the complete model is able to reduce the style loss in stylized images (e.g., the Starry Night, the Mosaic, the Composition VII, the La Muse styles), compared to the model w/o feature injection. 
These observations indicate that the adaptive feature injection effectively improves to keep the balance in content and style of stylized images. 

\begin{table}[tb]
\centering
\caption{Averages of $length$ (smaller is better) and $balance$ (larger is better) in the complete model (denoted by \textbf{complete}) and the model w/o feature injection (denoted by \textbf{w/o injection}).}
\vspace*{-0.5\baselineskip}
\label{tab:length_balance_comparison_detail}
\resizebox{1\linewidth}{!}{
\begin{tabular}{l@{\hspace*{0.2em}}|c@{\hspace*{0.35em}}c@{\hspace*{0.35em}}c@{\hspace*{0.35em}}c@{\hspace*{0.35em}}c@{\hspace*{0.35em}}}
\toprule[1pt]\midrule[0.3pt]
\multirow{2}{*}{\textbf{Style}} & \multicolumn{2}{c}{$length$ ($\Downarrow$)} & \quad & \multicolumn{2}{c}{$balance$ ($\Uparrow$)} \\ 
& \bf complete & \bf w/o injection & \quad & \bf complete & \bf w/o injection\\
\midrule
Starry Night & \bf{0.34} & 0.46 & \quad & \bf{2.12} & 1.35 \\
Mosaic & \bf{0.45} & 0.57 & \quad & \bf{1.80} & 1.03 \\
Composition VII & \bf{0.21} & 0.26 & \quad & \bf{3.61} & 3.21 \\
La Muse & 0.30 & \bf{0.27} & \quad & 1.72 & \bf{2.54} \\
The Wave & \bf{0.15} & 0.24 & \quad & \bf{5.39} & 3.15\\
Feathers & \bf{0.23} & 0.28 & \quad & \bf{3.71} & 3.20\\ 
\midrule
All together & \bf{0.28} & 0.35 & \quad & \bf{3.06} & 2.41 \\ 
\midrule[0.3pt]\bottomrule[1pt]
\end{tabular}
}
\end{table}

\subsubsection{Effectiveness of combination weight $\alpha$ and balance weight $\gamma$}


Here, we evaluate the necessity of combination weight $\alpha$ in Eq.~\eqref{eq:total-loss} and balance weight $\gamma$ in Eq.~\eqref{eq:balance_weight}.
In particular, we evaluate whether $\alpha$ plays the role of explicitly controlling the contribution ratio of the content and the style.

\begin{figure*}[tb]
	\centering
	\includegraphics[width=\linewidth]{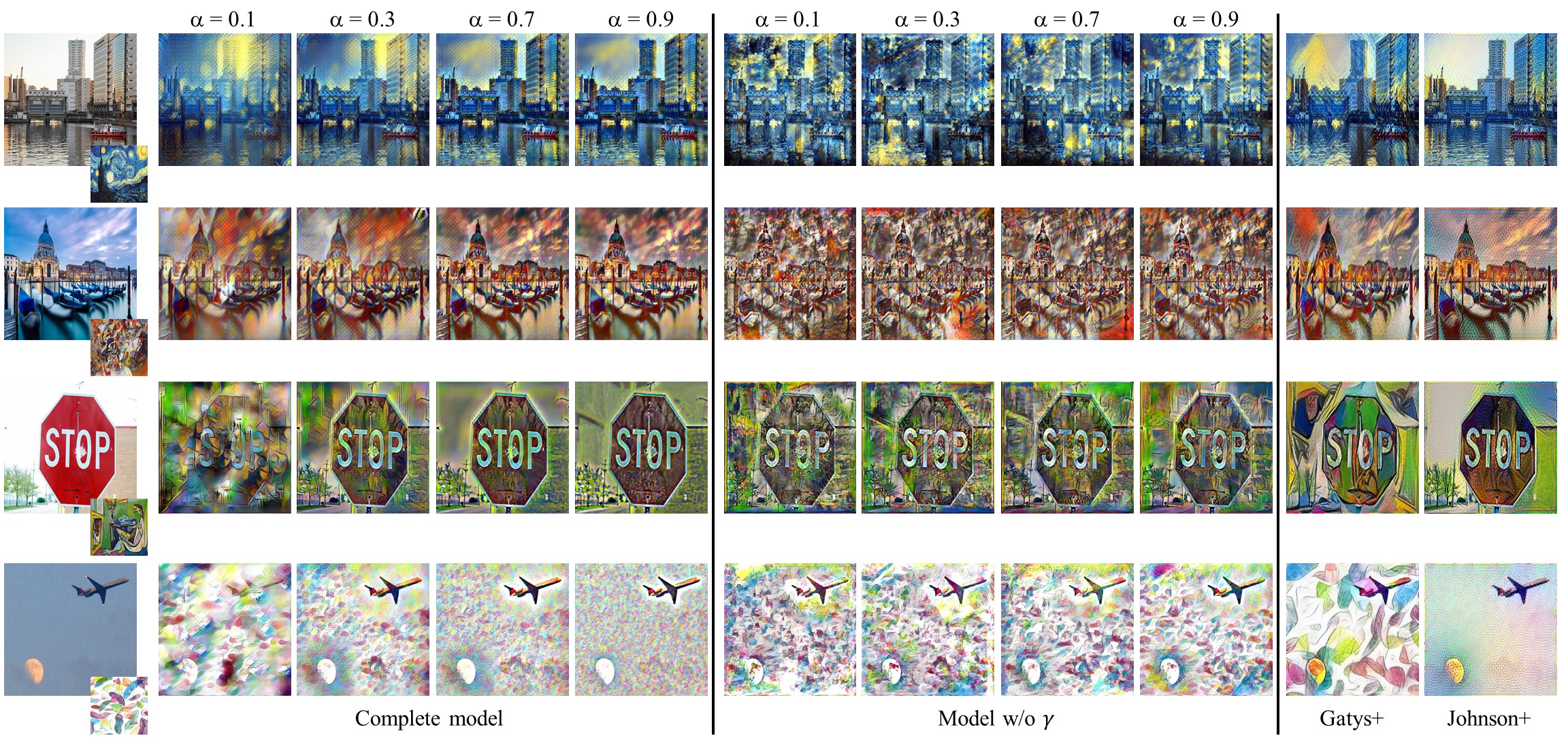}
    \vspace*{-1.0\baselineskip}
	\caption{Example of stylized images by changing $\alpha$ from 0.1 to 0.9. Left-most column: the content image (large) and the style image (small). From left to right: the stylized image using various $\alpha$. The last column shows results obtained by Gatys+~\cite{gatys2016image} and Johnson+~\cite{Johnson2016Perceptual} for the reference.} 
	\label{fig:change_alpha}
\end{figure*}

We generated stylized images using different values of $\alpha$: $\alpha = 0.1, 0.3, 0.7, 0.9$.
The results are illustrated in Fig.~\ref{fig:change_alpha} where the complete model denotes the model using $\alpha$ and $\gamma$ together while the model w/o $\gamma$ denotes the model using $\alpha$ only (i.e., $\gamma$ is disabled).
Ideally, for smaller $\alpha$, the style is more emphasized and results become more similar to those by Gatys+~\cite{gatys2016image}.
For larger $\alpha$, on the other hand, the content is more emphasized and results become more similar to those by Johnson+~\cite{Johnson2016Perceptual}.
We observe these in Fig.~\ref{fig:change_alpha} and see that $\alpha$ of the complete model indeed controls the contribution ratio of the content and the style as we expected.
However, we see that the model w/o $\gamma$ is not the case.
This observation suggests the necessity of both $\alpha$ and $\gamma$.





\section{Conclusion} \label{conclusion}

We presented an end-to-end two-stream network for balancing the content and style in stylized images. Our proposed method utilizes a deep FCN to preserve the semantic content and a shallow FCN to faithfully learn the style representation, whose outputs are adaptively feature injected and concatenated using the balance weight and fed into the decoder to generate stylized images. Our intensive experiments using six famous styles widely used in style transfer demonstrate the effectiveness of our proposed method against state-of-the-art methods in terms of balancing content and style. Furthermore, our proposed method outperforms the state-of-the-art methods in speed.

Our proposed method requires fine-tuning of parameters from an existing model to deal with different styles.  This limits the applicability of our proposed method to multi-style transfer.  
Extending our proposed method so that it can deal with a large style dataset such as Wikiart or unseen styles is left for future work.

\begin{figure}[tb]
	\centering
	\includegraphics[width=1\linewidth]{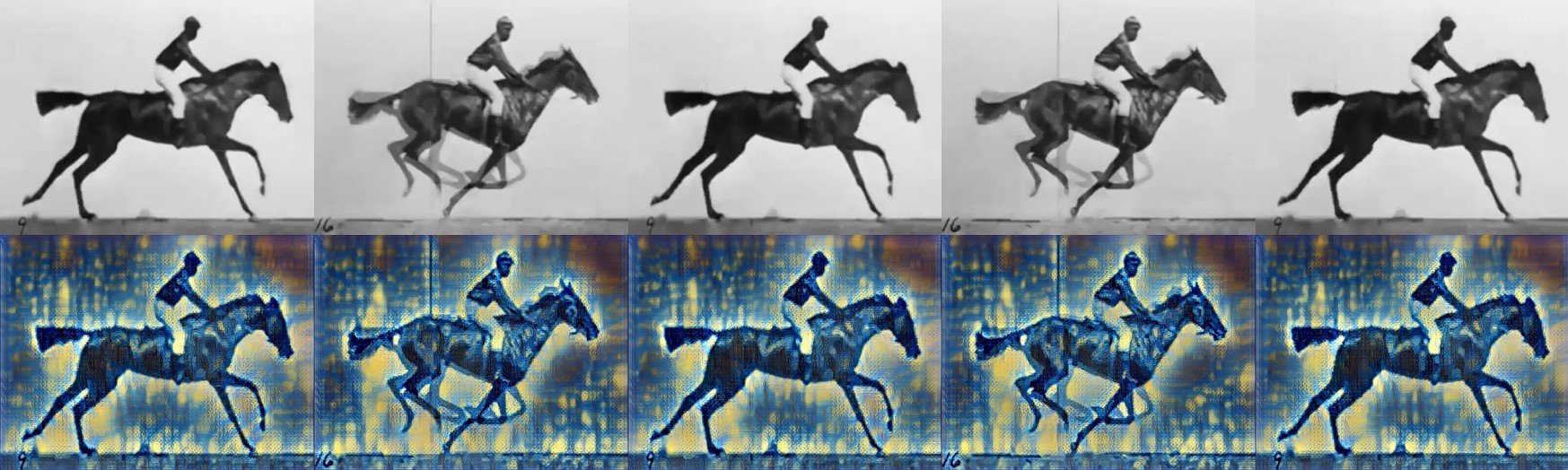}
    \vspace*{-1.0\baselineskip}
	\caption{Examples of stylized video in real-time using the "Starry night" style. We use the video of Eadweard Muybridge "The horse in motion" (1878) as the content input. Our model processes every frame independently without any post-processing. Video resolution is $480 \times 640$ at 30 FPS.} 
	\label{fig:video}
\end{figure}

As an extension of image style transfer, the real-time video stylization methods are currently proposed~\cite{Gao2018Reconet,Huang2017Realtime,Li2018Evolvement}. 
Since our proposed method runs fast, we believe that it can be useful for real-time video stylization. 
Though video stylization is out of the scope of this paper, we applied our method in the frame-by-frame manner to several videos for video stylization demonstration. 
Fig.~\ref{fig:video} shows some examples of stylized frames from a video. 
Our approach was able to stylize videos in real-time with the resolution $480 \times 640$ at 30 FPS or more. 
As we see, our method produces reasonable results for consecutive frames with varying appearance, meaning that the usage of our method for real-time video stylization is promising. 
We remark that we did not use either temporal regularization or post-processing.
Different from image style transfer, real-time video stylization needs to pay attentions to the temporal consistency among adjacent video frames.
Incorporating the temporal consistency into our method for real-time video stylization is left for our future work.

\section*{Acknowledgements}
This is a pre-print of an article published in Machine Vision and Applications. The final authenticated version is available online at: https://doi.org/10.1007/s00138-020-01086-1

This work was in part supported by JST CREST (Grant No. JPMJCR14D1).  
The authors are thankful to Dr. Trung-Nghia Le for his valuable comments on this work.

\bibliographystyle{IEEEtran}
\bibliography{references}

\begin{thebibliography}{10}
\providecommand{\url}[1]{#1}
\csname url@samestyle\endcsname
\providecommand{\newblock}{\relax}
\providecommand{\bibinfo}[2]{#2}
\providecommand{\BIBentrySTDinterwordspacing}{\spaceskip=0pt\relax}
\providecommand{\BIBentryALTinterwordstretchfactor}{4}
\providecommand{\BIBentryALTinterwordspacing}{\spaceskip=\fontdimen2\font plus
\BIBentryALTinterwordstretchfactor\fontdimen3\font minus
  \fontdimen4\font\relax}
\providecommand{\BIBforeignlanguage}[2]{{%
\expandafter\ifx\csname l@#1\endcsname\relax
\typeout{** WARNING: IEEEtran.bst: No hyphenation pattern has been}%
\typeout{** loaded for the language `#1'. Using the pattern for}%
\typeout{** the default language instead.}%
\else
\language=\csname l@#1\endcsname
\fi
#2}}
\providecommand{\BIBdecl}{\relax}
\BIBdecl

\bibitem{Johnson2016Perceptual}
J.~Johnson, A.~Alahi, and L.~Fei-Fei, ``Perceptual losses for real-time style
  transfer and super-resolution,'' in \emph{ECCV}, 2016.

\bibitem{huang2017adain}
X.~Huang and S.~Belongie, ``Arbitrary style transfer in real-time with adaptive
  instance normalization,'' in \emph{ICCV}, 2017.

\bibitem{gatys2016image}
L.~A. Gatys, A.~S. Ecker, and M.~Bethge, ``Image style transfer using
  convolutional neural networks,'' in \emph{CVPR}, 2016.

\bibitem{sheng2018avatar}
L.~Sheng, Z.~Lin, J.~Shao, and X.~Wang, ``Avatar-net: Multi-scale zero-shot
  style transfer by feature decoration,'' in \emph{CVPR}, 2018.

\bibitem{Chen}
T.~Q. Chen and M.~Schmidt, ``Fast patch-based style transfer of arbitrary
  style,'' in \emph{NIPS}, 2016.

\bibitem{Li2017Universal}
Y.~Li, C.~Fang, J.~Yang, Z.~Wang, X.~Lu, and M.-H. Yang, ``Universal style
  transfer via feature transforms,'' in \emph{NIPS}, 2017.

\bibitem{Kyprianidis2013}
J.~E. Kyprianidis, J.~Collomosse, T.~Wang, and T.~Isenberg, ``State of the
  "art": A taxonomy of artistic stylization techniques for images and video,''
  \emph{IEEE Transactions on Visualization and Computer Graphics}, 2013.

\bibitem{Ashikhmin2001}
M.~Ashikhmin, ``Synthesizing natural textures,'' in \emph{Symposium on
  Interactive 3D Graphics}, 2001.

\bibitem{Efros2001}
A.~A. Efros and W.~T. Freeman, ``Image quilting for texture synthesis and
  transfer,'' in \emph{SIGGRAPH}, 2001.

\bibitem{JingYFYS17}
Y.~Jing, Y.~Yang, Z.~Feng, J.~Ye, Y.~Yu, and M.~Song, ``Neural style transfer:
  A review,'' \emph{IEEE Transactions on Visualization and Computer Graphics},
  2019.

\bibitem{luan2017deep}
F.~Luan, S.~Paris, E.~Shechtman, and K.~Bala, ``Deep photo style transfer,'' in
  \emph{CVPR}, 2017.

\bibitem{mechrez2017photo}
R.~Mechrez, E.~Shechtman, and L.~Zelnik-Manor, ``Photorealistic style transfer
  with screened poisson equation,'' in \emph{BMVC}, 2017.

\bibitem{azadi2018multi}
S.~Azadi, M.~Fisher, V.~Kim, Z.~Wang, E.~Shechtman, and T.~Darrell,
  ``Multi-content gan for few-shot font style transfer,'' in \emph{CVPR}, 2018.

\bibitem{Dmytro2019Content}
D.~Kotovenko, A.~Sanakoyeu, P.~Ma, S.~Lang, and B.~Ommer, ``A content
  transformation block for image style transfer,'' in \emph{CVPR}, 2019.

\bibitem{Park2019Arbitrary}
D.~Y. Park and K.~H. Lee, ``Arbitrary style transfer with style-attentional
  networks,'' in \emph{CVPR}, 2019.

\bibitem{sanakoyeu2018styleaware}
A.~Sanakoyeu, D.~Kotovenko, S.~Lang, and B.~Ommer, ``A style-aware content loss
  for real-time hd style transfer,'' in \emph{ECCV}, 2018.

\bibitem{Wang2017CVPR}
X.~Wang, G.~Oxholm, D.~Zhang, and Y.-F. Wang, ``Multimodal transfer: A
  hierarchical deep convolutional neural network for fast artistic style
  transfer,'' in \emph{CVPR}, 2017.

\bibitem{zhang2018separating}
Y.~Zhang, Y.~Zhang, and W.~Cai, ``Separating style and content for generalized
  style transfer,'' in \emph{CVPR}, 2018.

\bibitem{li2018aclosed-form}
Y.~Li, M.-Y. Liu, X.~Li, M.-H. Yang, and J.~Kautz, ``A closed-form solution to
  photorealistic image stylization,'' in \emph{ECCV}, 2018.

\bibitem{duc2018balancing}
D.~M. Vo, T.~N. Le, and A.~Sugimoto, ``Balancing content and style with
  two-stream fcns for style transfer,'' in \emph{WACV}, 2018.

\bibitem{Heeger}
D.~J. Heeger and J.~R. Bergen, ``Pyramid-based texture analysis/synthesis,'' in
  \emph{SIGGRAPH}, 1995.

\bibitem{Li2017Laplacian}
S.~Li, X.~Xu, L.~Nie, and T.-S. Chua, ``Laplacian-steered neural style
  transfer,'' in \emph{ACM-MM}, 2017.

\bibitem{Ulyanov2016Texture}
D.~Ulyanov, V.~Lebedev, A.~Vedaldi, and V.~Lempitsky, ``Texture networks:
  Feed-forward synthesis of textures and stylized images,'' in \emph{ICML},
  2016.

\bibitem{Radford2015Unsupervised}
A.~Radford, L.~Metz, and S.~Chintala, ``Unsupervised representation learning
  with deep convolutional generative adversarial networks,'' in \emph{ICLR},
  2016.

\bibitem{dumoulin2017learned}
V.~Dumoulin, J.~Shlens, and M.~Kudlur, ``A learned representation for artistic
  style,'' in \emph{ICLR}, 2017.

\bibitem{UlyanovVL16}
D.~Ulyanov, A.~Vedaldi, and V.~S. Lempitsky, ``Instance normalization: The
  missing ingredient for fast stylization,'' in \emph{ICML}, 2016.

\bibitem{ian2014generative}
I.~Goodfellow, J.~Pouget-Abadie, M.~Mirza, B.~Xu, D.~Warde-Farley, S.~Ozair,
  A.~Courville, and Y.~Bengio, ``Generative adversarial nets,'' in \emph{NIPS},
  2014.

\bibitem{Li2016Precomputed}
C.~Li and M.~Wand, ``Precomputed real-time texture synthesis with markovian
  generative adversarial networks,'' in \emph{ECCV}, 2016.

\bibitem{SimonyanZ14a}
K.~Simonyan and A.~Zisserman, ``Very deep convolutional networks for
  large-scale image recognition,'' in \emph{ICLR}, 2015.

\bibitem{RussakovskyDSKS15}
O.~Russakovsky, J.~Deng, H.~Su, J.~Krause, S.~Satheesh, S.~Ma, Z.~Huang,
  A.~Karpathy, A.~Khosla, M.~S. Bernstein, A.~C. Berg, and F.~Li, ``Imagenet
  large scale visual recognition challenge,'' \emph{International Journal of
  Computer Vision}, 2015.

\bibitem{Nair2010}
V.~Nair and G.~E. Hinton, ``Rectified linear units improve restricted boltzmann
  machines,'' in \emph{ICML}, 2010.

\bibitem{MahendranV15understanding}
A.~Mahendran and A.~Vedaldi, ``Understanding deep image representations by
  inverting them,'' in \emph{CVPR}, 2015.

\bibitem{Kingma2014}
D.~P. Kingma and J.~Ba, ``Adam: {A} method for stochastic optimization,'' in
  \emph{ICLR}, 2015.

\bibitem{He2015}
K.~He, X.~Zhang, S.~Ren, and J.~Sun, ``Deep residual learning for image
  recognition,'' in \emph{CVPR}, 2016.

\bibitem{Ioffe2015}
S.~Ioffe and C.~Szegedy, ``Batch normalization: Accelerating deep network
  training by reducing internal covariate shift,'' in \emph{ICML}, 2015.

\bibitem{Horn2012}
R.~A. Horn and C.~R. Johnson, \emph{Matrix Analysis}, 2nd~ed.\hskip 1em plus
  0.5em minus 0.4em\relax New York, NY, USA: Cambridge University Press, 2012.

\bibitem{Lin2014}
T.-Y. Lin, M.~Maire, S.~Belongie, J.~Hays, P.~Perona, D.~Ramanan, P.~Dollár,
  and C.~L. Zitnick, ``Microsoft coco: Common objects in context,'' in
  \emph{ECCV}, 2014.

\bibitem{Gao2018Reconet}
C.~Gao, D.~Gu, F.~Zhang, and Y.~Yu, ``Reconet: Real-time coherent video style
  transfer network,'' in \emph{ACCV}, 2018.

\bibitem{Huang2017Realtime}
H.~Huang, H.~Wang, W.~Luo, L.~Ma, W.~Jiang, X.~Zhu, Z.~Li, and W.~Liu,
  ``Real-time neural style transfer for videos,'' in \emph{CVPR}, 2017.

\bibitem{Li2018Evolvement}
W.~Li, L.~Wen, X.~Bian, and S.~Lyu, ``Evolvement constrained adversarial
  learning for video style transfer,'' in \emph{ACCV}, 2018.

\end{thebibliography}

\end{document}